\documentclass[10pt,twocolumn,letterpaper]{article}

\usepackage[final]{cvpr}

\usepackage{times}
\usepackage{epsfig}
\usepackage{graphicx}
\usepackage{amsmath}
\usepackage{amssymb}
\usepackage[percent]{overpic}
\usepackage{multirow}
\usepackage{cuted}
\usepackage{flushend}
\usepackage[hypcap=false]{caption}
\usepackage{filecontents}
\usepackage{pgfplots}
\usepackage{pgfplotstable}

\usepackage[pagebackref,breaklinks,colorlinks]{hyperref}

\usepackage[capitalize]{cleveref}
\crefname{section}{Sec.}{Secs.}
\Crefname{section}{Section}{Sections}
\Crefname{table}{Table}{Tables}
\crefname{table}{Tab.}{Tabs.}

\def\cvprPaperID{xxxxx} 
\def\confName{CVPR}
\def\confYear{2022}

\def\eg{\emph{e.g.}}

\def\etal{\emph{et al.}}

\newcommand{\beginsupplement}{%
        \setcounter{section}{0}
        \renewcommand{\thesection}{\Alph{section}}%
        \setcounter{subsection}{0}
        \renewcommand{\thesubsection}{\Alph{section}.\arabic{subsection}}
        \setcounter{table}{0}
        \renewcommand{\thetable}{\Alph{section}\arabic{table}}%
        \setcounter{figure}{0}
        \renewcommand{\thefigure}{\Alph{section}\arabic{figure}}%
}

\def\poincare/{Poincar\'e}

\def\eg{\textit{e.g.}}
\def\etal{\textit{et al.}~}

\def\mobius/{M{\"o}bius}

\begin{document}


\DeclareGraphicsExtensions{.jpg,.png,.pdf}





\title{Neural Convolutional Surfaces}

\author{
Luca Morreale$^{1}$\thanks{Partially worked on the project during internship at Adobe Research.}
\quad\;
Noam Aigerman$^{2}$
\quad\;
Paul Guerrero$^{2}$
\quad\;
Vladimir G. Kim$^{2}$
\quad\;
Niloy J. Mitra$^{1,2}$
\\
$^{1}$University College London
\quad\quad
$^{2}$Adobe Research
}

\maketitle

\begin{strip}
\centering
\includegraphics[width=\textwidth]{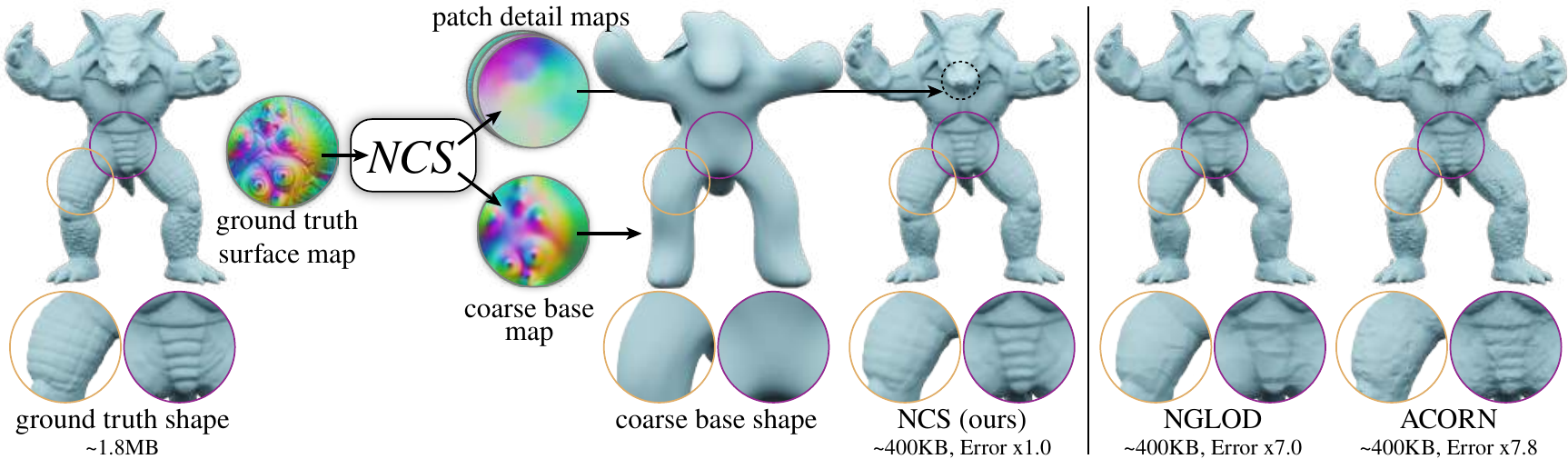}
\captionof{figure}{\label{fig:teaser}
Neural Convolutional Surfaces (NCS) can faithfully represent a given ground-truth shape while disentangling coarse geometry from fine details, leading to a highly-accurate representation of the shape. Compared to other state-of-the-art methods neuralLOD~\cite{takikawa2021neural} and Acorn~\cite{martel2021acorn}, NCS achieves significantly more accurate results for the same memory footprint. }
\end{strip}

\begin{abstract}

This work is concerned with a representation of shapes that disentangles fine, local and possibly repeating geometry, from global, coarse structures. Achieving such disentanglement leads to two unrelated advantages: i) a significant compression in the number of parameters required to represent a given geometry; ii) the ability to manipulate either global geometry, or local details, without harming the other. At the core of our approach lies a novel pipeline and neural architecture, which are optimized to represent one specific atlas, representing one 3D surface. Our pipeline and architecture are designed so that disentanglement of global geometry from local details is accomplished through optimization, in a completely unsupervised manner. We show that this approach achieves better neural shape compression than the state of the art, as well as enabling manipulation and transfer of shape details. Project page 
\url{http://geometry.cs.ucl.ac.uk/projects/2022/cnnmaps/}.

\end{abstract}


\section{Introduction}

Triangle meshes have been the most popular representation across much of geometry processing since its early stages, however research has been devoted to devising novel representations of geometry to circumvent many of the shortcoming of triangular meshes. Lately, the rising prominence of deep learning has lead researchers to investigate ways to represent shapes via neural networks. While the immediate use of neural networks in this context is to represent entire shape spaces by using the same set of weights to decode any shape from a shared latent space, other methods use a shape-specific set of weights to represent a specific instance. This approach captures geometric detail efficiently and accurately and creates outputs that are on par with existing 3D models, while holding novel properties not attainable with surface meshes, such as differentiability. These neural representations for shape instances were demonstrated to be useful in geometry processing applications such as efficient rendering~\cite{takikawa2021neural}, level of details~\cite{martel2021acorn}, surface parameterization, and inter-surface mapping~\cite{morreale2021neural}.

The choice of the shape representation, and of the neural network's architecture, plays a critical role in how efficiently the capacity of the network is utilized. Existing representations usually use MLPs to model the shape as a function that maps points either from a 2D atlas to the surface~\cite{morreale2021neural} or points in a 3D volume to an implicit function such as a distance field~\cite{park2019deepsdf}. The disadvantage of these architectures is that they entangle geometric details and overall shape structure, and do not have a natural mechanism to reuse the network weights to represent repeating local details, as Convolution Neural Networks~(CNNs) achieve on images. Some methods indeed opt to use 2D images to represent geometry \cite{Sinha2016DeepL3}, however those exhibit finite resolution and hence cannot model surfaces with details in sub-pixel resolution.
Alternatively, instead of a single global MLP, some prior techniques leverage repetitions by breaking the shape into smaller 3D voxels, each represented by an SDF function~\cite{martel2021acorn}, however, these representations do not account for the fact that surface details are usually aligned with the surface, and thus, are less effective at representing local geometric textures that flow with the shape.

In this paper, we set to define a novel representation that achieves separation of local geometric details (``texture") from the global coarse geometry of the model, and thus leads to the reuse of network weights for repeating patterns that change their orientation with the surface. We achieve this by considering the standard atlas-based representation as in \cite{groueix2018papier,morreale2021neural}, but encode a surface as combination of a \textit{coarse surface}, defining the general, coarse structure of the shape, represented via an MLP, along with an associated fine \textit{detail map}, which adds geometric texture on top, represented via a CNN, which defines a continuous map of offsets. The geometric details are added to the coarse geometry either along its normal directions, or as general displacement vectors. Since the local displacement details are expressed with convolutional kernels, they can effectively be reused across similar regions of the surface. We call this hybrid representation \textit{neural convolutional surfaces}.

This novel architecture enables the network to disentangle the fine CNN representation from the coarse MLP representation, in a completely \emph{unsupervised} manner, i.e., without
the need to supervise the split explicitly during fitting. We show that
the inductive bias in our designed architecture leads to automatic separation of the shapes into coarse base shapes and reusable convolutional details, see Figure~\ref{fig:teaser}.

We evaluate our method on a range of complex surfaces and explore the associated tradeoff between representation quality and model complexity. We compare against a set of state-of-the-art alternatives (e.g., NeuralLod~\cite{takikawa2021neural}, ACORN~\cite{martel2021acorn}, Neural Surface Maps~\cite{morreale2021neural}) and demonstrate that our model achieves better accuracy at a fraction of the model-complexity -- between 1\% to 10\% parameters.
Additionally, we demonstrate that the convolutional aspect of the representation makes it \emph{interpretable}, leading to applications including detail modification within individual shapes and details transfer across different models -- see Figure \ref{fig:interp}.

\section{Related Works}

Our method follows a long line of works on using neural shape representation, as well as the more specific recent trend of representing a single shape via a neural network. We also incorporate concepts on level-of-detail representation of geometry. Next, we review the existing literature on all these fields.

\paragraph{Neural representation of shape spaces.}
A large number of generative neural representations for 3D shapes have been proposed in recent years, such as voxel grids~\cite{liu2018voxelgan,GirdharFRG16,BrockLRW16,dai2017complete}, point clouds~\cite{achlioptas2018latent_pc,Su2017PointGen},   meshes~\cite{dai2019scan2mesh}, or unions of deformable primitives~\cite{genova2019}. With these representations, the number or scale of the discrete elements has a great effect on the method's ability to represent fine details. To tackle this challenge, some methods model shapes as neural functions, allowing the network to optimize how to allocate its capacity:
implicit models represent the shape as a map from a volume to a signed distance field~\cite{park2019deepsdf} or occupancy~\cite{occupancy_nets_2019} values. These can be further improved by enforcing satisfaction of the Eikonal equation ~\cite{gropp_implicit_2020,atzmon_sal_2020,atzmon_sal_2019} or using an intermediate meta-network for faster reconstruction~\cite{littwin_deep_2019}. Since most traditional computer-graphics pipelines require surface models, atlas-based representations offer another prominent alternative. These techniques model a shape as an atlas, i.e., a function that maps 2D points to positions in 3D~\cite{groueix2018papier,yang_foldingnet_2018}. Improved versions of these methods include adding optimizing for low-distortion atlases~\cite{bednarik_shape_2019}, learning task-specific geometry of 2D domain~\cite{deprelle_learning_2019}, or forcing the surface to agree with an implicit function~\cite{Poursaeed20a}. \cite{Sinha2016DeepL3} use Geometry Images \cite{gu_geom} and encode geometry as 2D images in order to perform deep learning tasks such as segmentation.

\begin{figure*}[t]
    \includegraphics[width=\textwidth]{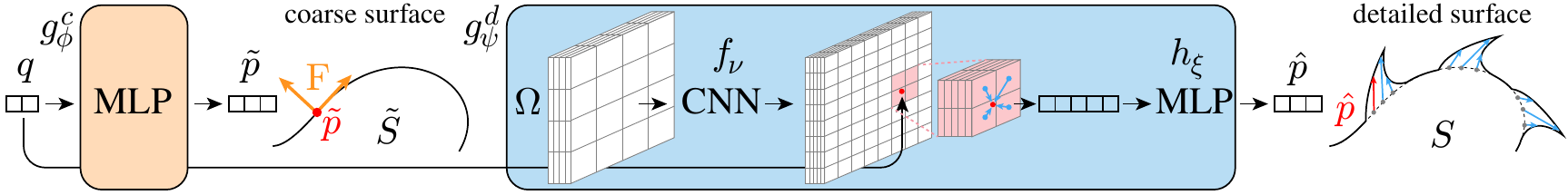}
        
    
    \caption{ \label{fig:architecture}
        \textbf{Overview of Neural Convolutional Surfaces.} Surfaces $S$ are represented by two models, a coarse model $g_\phi^c$ that encodes a coarse version $\tilde{S}$ of the surface and allows computing a local reference frame $\mathbf{F}$, and fine model $g_\psi^d$ that encodes geometric detail as offsets $\hat{p}$ from the coarse surface, in coordinates of the local reference frame.
    }
\end{figure*}

\paragraph{Fitting networks to shapes.}
While the previously-mentioned works are concerned with training networks to represent an arbitrary shape out of a collection, similar techniques can be employed to represent one, specific shape via one network, fitted to that specific shape,  by optimizing the network's weights on that single example. This has several advantages; for example, Neural Surface Maps~\cite{morreale2021neural} use a fully connected neural network to represent a parameterized shape as an $R^2 \rightarrow R^3$ map and show that the achieved reconstruction is much more accurate and can represent fine details than a network trained to reconstruct multiple shapes. Furthermore, one can concatenate these maps and optimize them to achieve low-distortion surface-to-surface maps. Representing a shape via a neural network enables compression, of, e.g., implicit surfaces ~\cite{davies2020overfit,takikawa2021neural}. One can also optimize from indirect observations, such as multi-view appearance~\cite{sitzmann_implicit_2020,mildenhall_nerf_2020}. 

Many of these works are concerned with preserving details. ACORN~\cite{martel2021acorn} solves an integer-program to optimally subdivide an SDF volume to allocate more parameters to the regions holding more details. Yifan \etal~propose siren-based implicit displacement fields~\cite{yifan2021geometryconsistent} restricted to the normal direction, which separate geometric details from the coarse shape structure. In our work, we use convolutional filters applied on the 2D domain to model displacement from the atlas-based surface representation. Using convolutional filters enables us to rely on repeating signals (which could be encoded in those filters applied throughout the surface). By modeling our displacements relative to the parameterized surface atlas, we can learn also learn geometric textures that are aligned with the surface.

\paragraph{Level-of-detail in geometry processing.}
Many prior optimization techniques focus on simplifying meshes with high polygon count. Different simplification operators~\cite{Schroeder1997,Tan1997,garland1997} with geometry-based and appearance-based~\cite{Luebke1997} objective functions (see ~\cite{Luebke2002} for a thorough overview). In this work, instead of simplifying an existing mesh, we propose to use a neural network to represent the surface. One can control the amount of geometric detail by simply changing the capacity of the neural network. Our representation can also effectively represent geometric texture by using neural convolutional surfaces that utilize shared kernels for repeating patterns.

\section{Neural Convolutional Surfaces} \label{sec:method}

Our goal is to represent a 3D surface $S \subset \mathbb{R}^3$ with a neural network $g_\theta$, so that the parameters $\theta$ compactly encode the surface. 
%
We assume to have computed a bijective parameterization of the surface into the unit circle in 2D (we use SLIM~\cite{rabinovich2017scalable} in all our experiments). We consider this parameterization as a map from the unit disk to the surface $D : \{ \mathbf{x} \in \mathbb{R}^2: \| \mathbf{x}\| \le 1\} \rightarrow S$, mapping a 2D point $q$ to a 3D point $p \in S$ on the surface. Following \cite{morreale2021neural}, we fit the network $g_\theta$ to approximate this function: $g_\theta(q) \approx s(q)$. 

We propose \emph{Neural Convolutional Surfaces} (NCS) as an accurate and compact neural approach to model $s$. The NCS $g_\theta$ consists of two modules: (i)~a coarse module,  $g^c_{\phi}$, based on a standard MLP-based model, which aims to approximate the coarse shape 
of the surface; and (ii)~a fine module, $g^d_{\psi}$, which adds detailed displacements 
to the coarse surface. Internally, this fine module is comprised of a CNN component generating a grid of codes, which are interpolated and fed to an MLP in order to get the final fine offset vector. These two modules are then added to get the final map from 2D to 3D,
\begin{equation}
g_\theta(q) := g^c_{\phi}(q)\ +\ \mathbf{F}_{g^c_{\phi}}(q) \hspace{3pt} g^d_{\psi}(q),
\end{equation}
where $\mathbf{F}_{g^c_{\phi}}$ is a rotation to the local reference frame at a coarse surface location and $\theta = (\phi, \psi)$.
%
Please refer to Figure~\ref{fig:architecture} for an overview.
These two models are trained \textit{jointly}, with only the target mapping $s$ as supervision. Intuitively, the coarse-fine separation allows the fine model $g^d_\psi$ to expend capacity only on the high-frequency geometric texture of the surface. Repeating structures in this geometric texture can be modelled efficiently by the shared weights of the convolution kernels.

 We describe the coarse model next in Section~\ref{sec:coarse_model}, the fine model in Section~\ref{sec:fine_model}, the local reference frame in Section~\ref{sec:ref_frame}, and the overall training setup in Section~\ref{sec:training}.

\subsection{Coarse Model}
\label{sec:coarse_model}
The coarse model represents a smooth, coarse version of the ground truth surface $S$, which serves as a basis, by providing a well defined local coordinate frame at any surface point, for applying the detailed geometric texture predicted by the fine model.

Our coarse model approximates a coarse version of $s$ with a low-capacity MLP, which takes as input a 2D coordinate $q$ in the unit disk and outputs the corresponding point $\tilde{p}$ on the coarse surface:
$\tilde{p} = g^c_\phi(q)$. The limited capacity of the MLP ensures that only a coarse approximation of the surface is modelled, while the MLP's architecture enforces smoothness of the coarse surface, thereby delegating reconstruction of sharp, fine, and repeating structures to the fine model. Note that we do \textit{not} use intermediate supervision for the coarse surface $\tilde{p}$. 

\subsection{Fine Model}
\label{sec:fine_model}
The fine model represents a detailed geometric texture that is applied to the coarse surface $g^c_\phi$. The fine model is designed to first generate a high-resolution 2D grid of codes, and then interpolate the codes and map the interpolated code to 3D displacement vectors via a small MLP. This is facilitated by three components: (i)~we keep a low-resolution input grid of learned features (i.e., a low-resolution 2D image with multiple channels)  $\Omega \in \mathbb{R}^{D_0 \times H_0 \times W_0}$; (ii)~a CNN $f_\nu$ transforms and up-samples(2x per layer) the feature map into a high-resolution 2D grid of codes; and (iii)~for a given 2D query point $q$ that falls into a grid cell, the 4 grid codes at that cell's corners are interpolated, and a small 2-layer MLP $h_\xi$ finally maps the interpolated code into a local displacement vector:
\begin{equation}
    g^d_\psi(q) = h_\xi\bigl(f_\nu(\Omega)|_q\bigr),
    \label{eq:fine_model_unpatched}
\end{equation}
where $X|_q$ denotes bilinear interpolation of the image $X$ at $q$ (assuming pixel coordinates in $[-1,1]^2$) and $\psi = (\Omega, \nu, \xi)$. Intuitively, the feature map $\Omega$ stores coarse information about the geometric surface details that is refined by the CNN $f_\nu$, introducing learned priors stored in the shared CNN kernels. The interpolation for a given sample $q$ is performed in feature space as opposed to 3D space, and followed by a small MLP to allow for complex non-linear interpolating surfaces between the pixels of the CNN output. For details of the model architectures, please refer to the supplementary material. 

\paragraph{Patches.} Directly discretizing the full parameter domain of $s$ on a grid has two drawbacks: (i)~The resolution of the pixel grid processed by the CNN would need to be very large to accurately model small geometric detail; and (ii)~the initial mapping $s$ may exhibit significant area distortion, i.e., there can be a large difference between scale factors in different regions of the mapping, making a single global resolution inefficient.

To avoid these problems, we split the surface $S$ into small overlapping patches $R_0, \dots, R_m$ with each having a separate local parameterization $r_i: [-1,1]^2 \rightarrow R_i$. Since each patch only covers a small region of the surface, the distortion within each individual mapping is small, and the resolution of the pixel grid in each patch can be lower, without compromising geometric detail. Further, since we assume our fine model is CNN-based, we can learn and reuse the same CNN kernels for each one of our patches without harming our goal of training the fine model to represent repeated geometry, now simply split into different patches.

\begin{figure}[t]
    \centering
    \setlength\tabcolsep{1.5pt} 
    \begin{tabular}{cccc}
        \includegraphics[width=0.32\columnwidth]{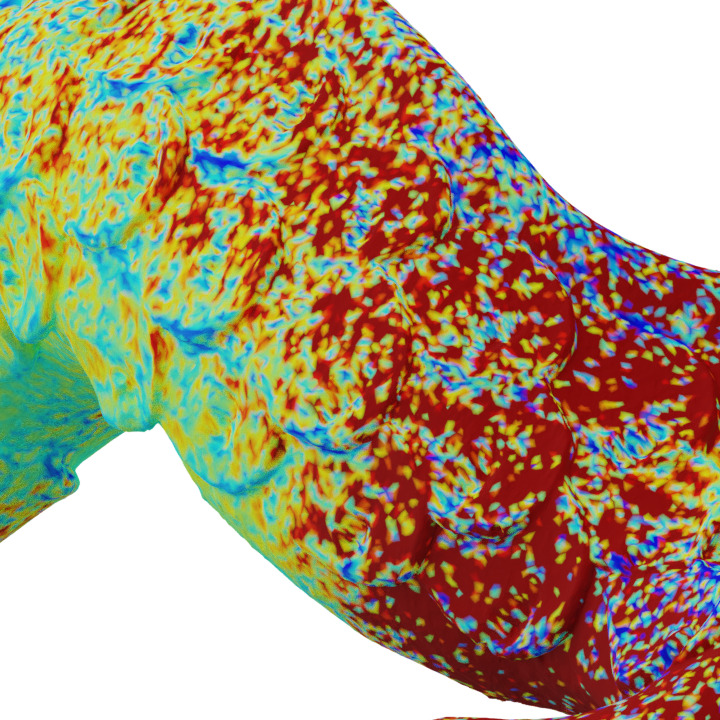} &
        \includegraphics[width=0.32\columnwidth]{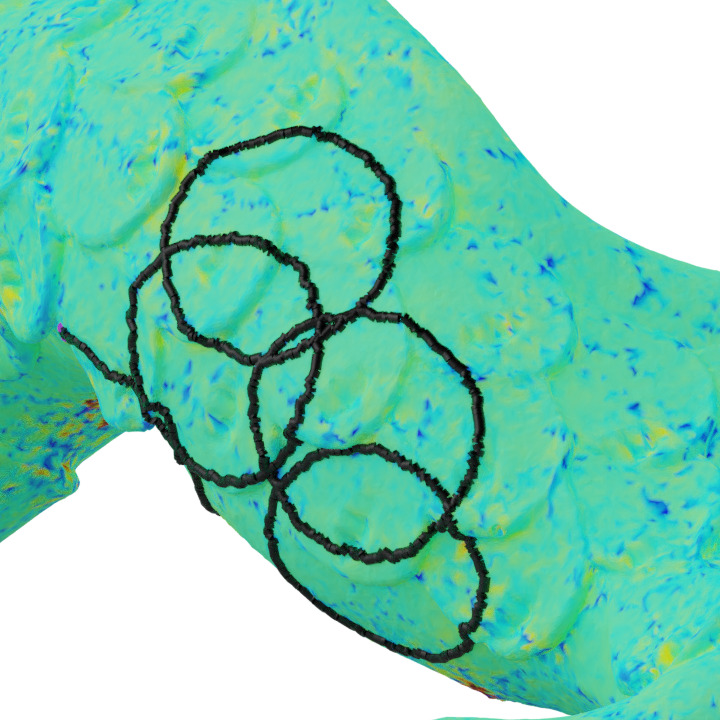} &
        \includegraphics[width=0.32\columnwidth]{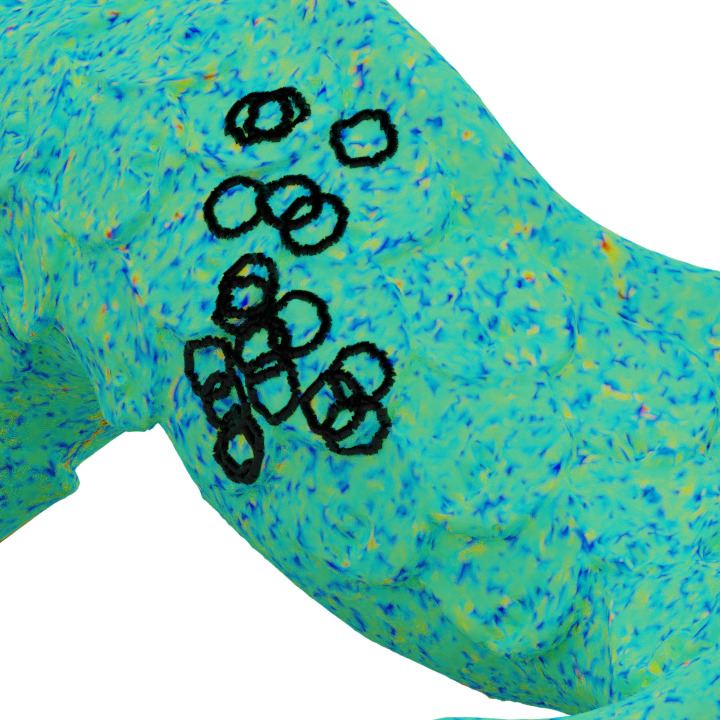} &
        \includegraphics[height=2.7cm]{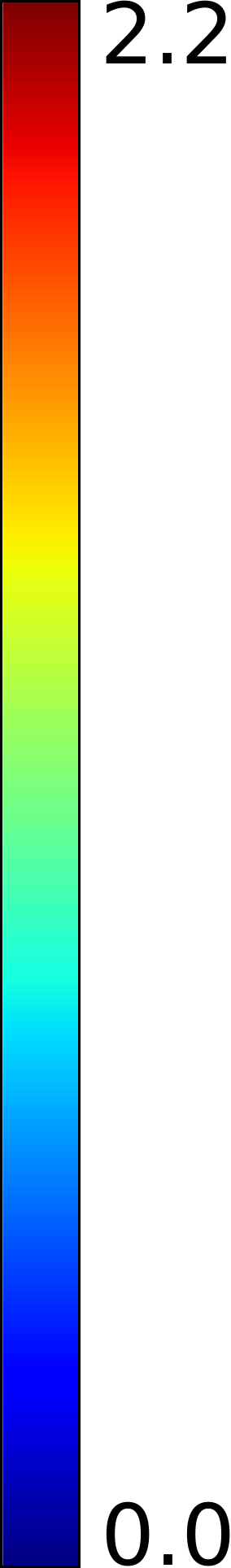} \\
        \textbf{(a)} 1 patch & \textbf{(b)} 700 patches & \textbf{(c)} $>$2000 patches \\
    \end{tabular}
    \caption{\textbf{Effect of number of patches.} Representing a shape with a single patch, results in high distortion in the underlying paramaterization
    (red in (a)). Decomposing the base domain into many small patches~(c) leads to much lower per-patch distortion. However, this comes at the cost of  a much increased memory budget to represent the shape. Medium size patches~(b) strikes a balance by reducing the per-patch distortion while still being light in the final memory requirement. 
    }
    \label{fig:effectOfPatchSize}
\end{figure}

\paragraph{Patch-based model.} Once we decompose the input into multiple patches, Equation~\ref{eq:fine_model_unpatched} can be generalized as:
\begin{gather}
    \label{eq:fine_model}
    g^d_\psi(q) = \frac{1}{\sum_i w_i(q)} \sum_i w_i(q)\ h_\xi\bigl(f_\nu(\Omega_i)|_{l_i(q)} \bigr)\\
    \text{with } w_i(q) = \max\bigl(0, -d^b_i\bigl(l_i(q)\bigr)\bigr), \nonumber
\end{gather}
where $l_i(q)$ maps the global parameters $q$ to the local patch parameterization $r_i$, and the contribution of overlapping patches to a point $q$ is weighed as a function of the signed distance $d^b_i$ from the boundary of the patch to $l_i(q)$ in the parameter domain of the patch. We train the same MLP and CNN (with shared weights $\nu, \xi$) over all patches, thereby encouraging the CNN to reuse filters across patches; the only different parameter between different patches is the coarse input feature grid map  $\Omega_i$ that is assigned to each patch. 




\subsection{Local Reference Frame}
\label{sec:ref_frame}
The output of the fine model $g^d_\psi(q)$ is a displacement vector $\hat{p}$ of the coarse surface at $\tilde{p} = g^c_\phi(q)$. Naively adding $\tilde{p} + \hat{p}$ would render the displacements sensitive to the local orientation of the coarse surface. Hence, to encourage  consistency between the displacements, we define them in a local coordinate frame $\mathbf{F}_{g^c_{\phi}}$ aligned to the tangent space of the coarse surface, measured via  the Jacobian of the coarse surface mapping, as:
\begin{equation}
\mathbf{J}^c :=  \left[ \mathbf{J}^c_u, \mathbf{J}^c_v \right] = 
\left[\frac{\partial g^c_{\phi}}{\partial q_u}, \frac{\partial g^c_{\phi}}{\partial q_v}\right],
\end{equation}
where $q_u$ and $q_v$ are the two coordinates of the global parameterization. The local coordinate frame as a function of $q$ is then defined as:
\begin{gather}
\mathbf{F}_{g^c_{\phi}} := \left[n,\ \mathbf{J}^c_u,\ n \wedge \mathbf{J}^c_u \right] \text{ with } n = \mathbf{J}^c_u \wedge \mathbf{J}^c_v,
\end{gather}
where $n$ returns the normal of the coarse surface and $\wedge$ denotes the cross product. Note that, although not shown in the expressions above, each of the axis vectors are normalized to be of unit length.

\subsection{Training}
\label{sec:training}
As discussed earlier, the coarse and fine modules together define a neural network mapping 2D points to 3D. We fit this combined map to the ground truth surface mapping $s$ via an L2 loss:
\begin{equation}
\mathcal{L}_\text{joint} = \int_{Q_S} \| g_\theta(q) - s(q) \|_2^2 \ dq,
\end{equation}
where $Q_S$ is the subregion of the global parameter domain $Q$ that maps to the surface $S$.

\begin{figure*}[t]
    \centering
    \setlength\tabcolsep{1.0pt} 
    \begin{tabular}{cccc}

        \includegraphics[width=0.46\columnwidth]{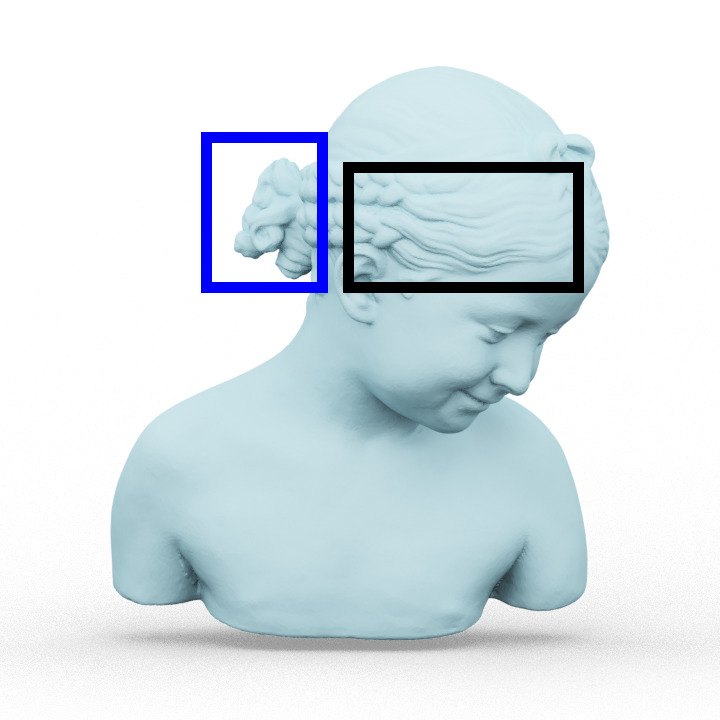} &
        \includegraphics[width=0.46\columnwidth]{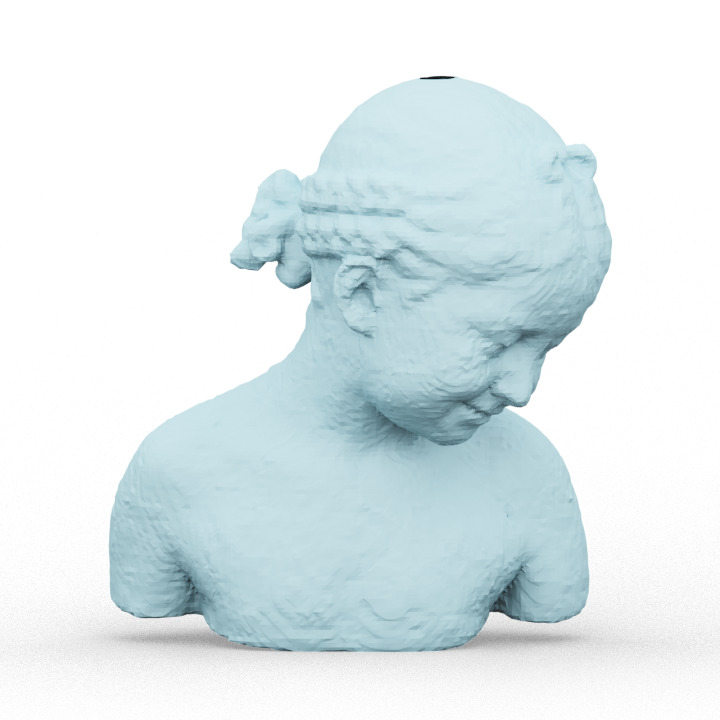} &
        \includegraphics[width=0.46\columnwidth]{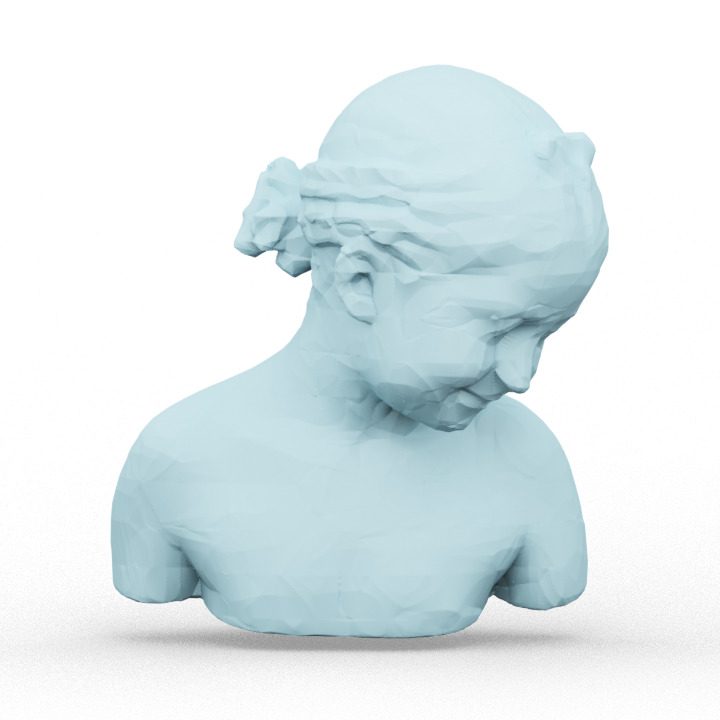} &
        \includegraphics[width=0.46\columnwidth]{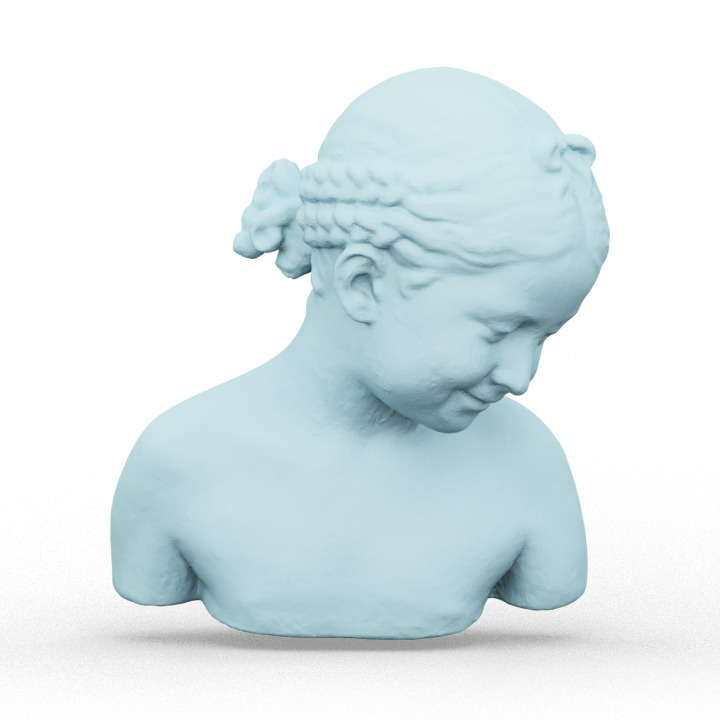} \\

        \includegraphics[width=0.46\columnwidth]{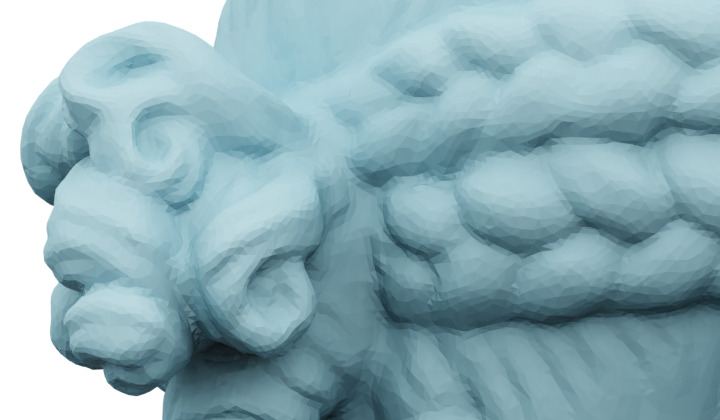} &
        \includegraphics[width=0.46\columnwidth]{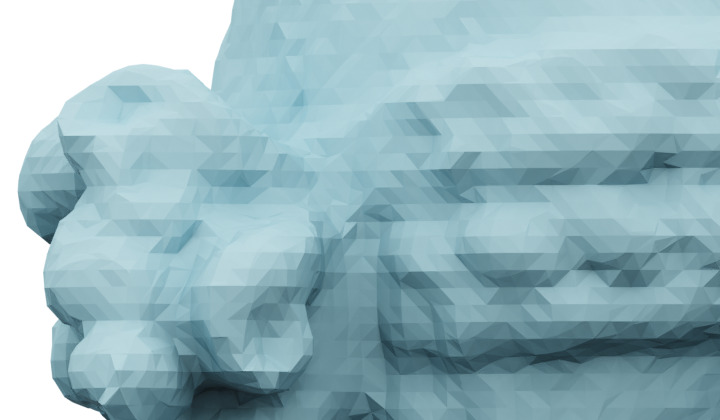} &
        \includegraphics[width=0.46\columnwidth]{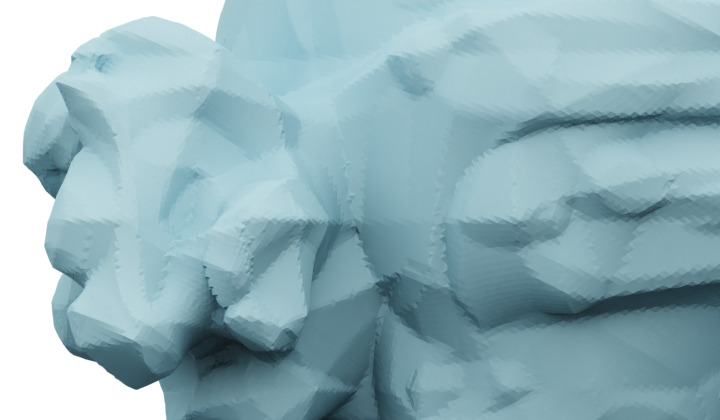} &
        \includegraphics[width=0.46\columnwidth]{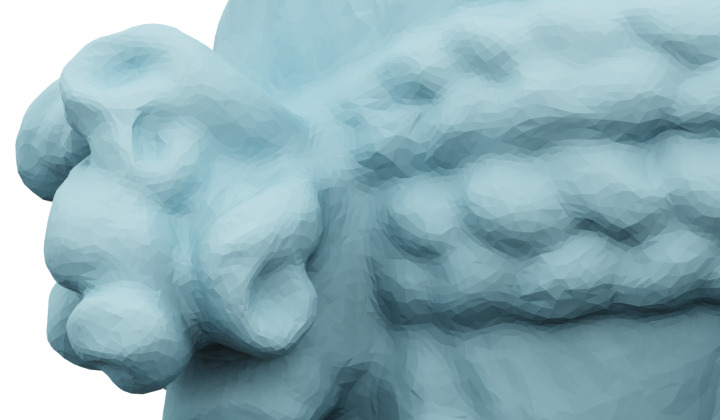} \\
        
        \includegraphics[width=0.46\columnwidth]{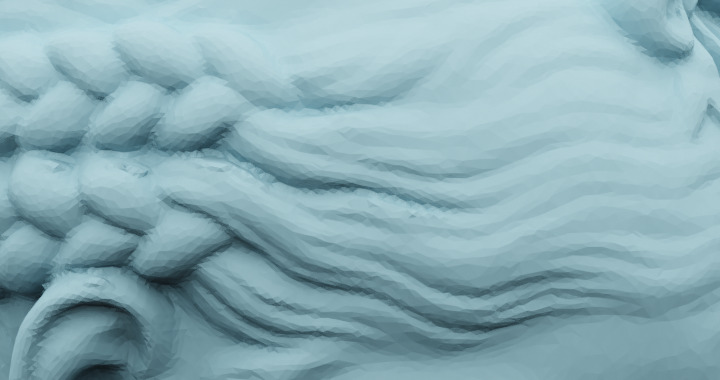} &
        \includegraphics[width=0.46\columnwidth]{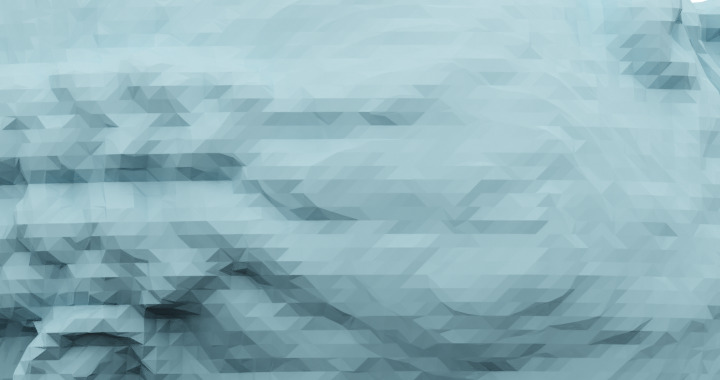} &
        \includegraphics[width=0.46\columnwidth]{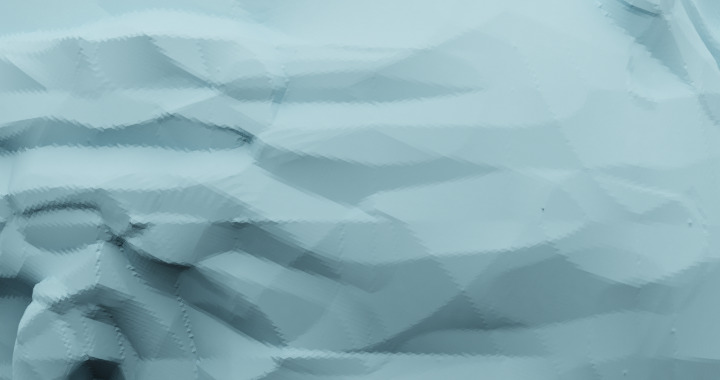} &
        \includegraphics[width=0.46\columnwidth]{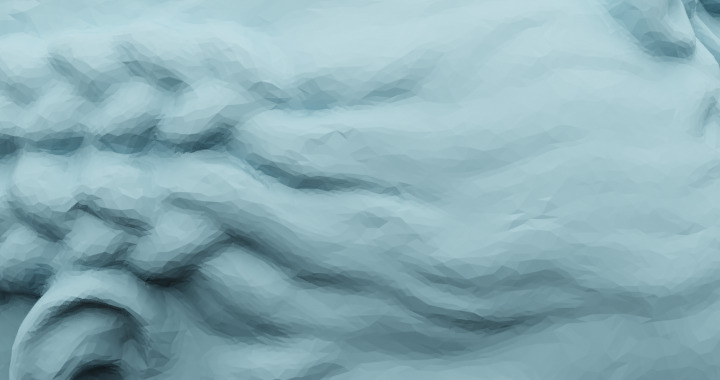} \\
        
        \includegraphics[width=0.46\columnwidth]{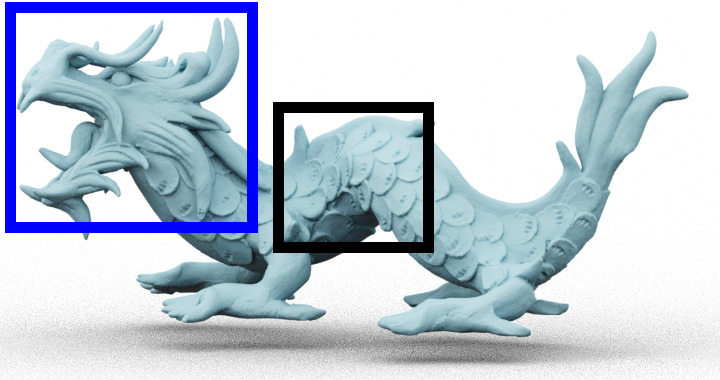} &
        \includegraphics[width=0.46\columnwidth]{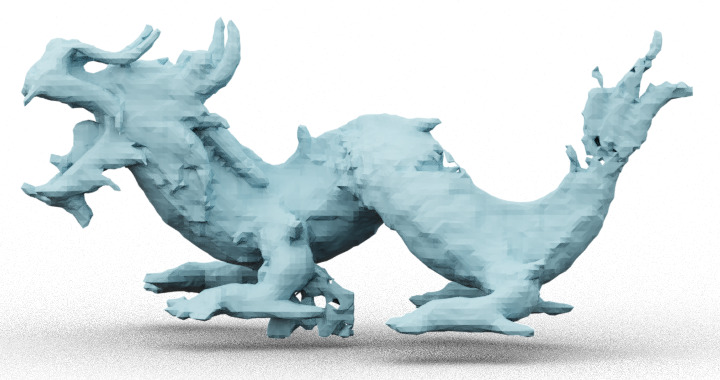} &
        \includegraphics[width=0.46\columnwidth]{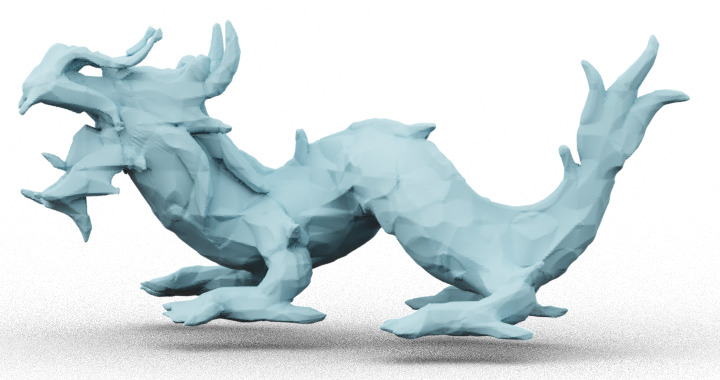} &
        \includegraphics[width=0.46\columnwidth]{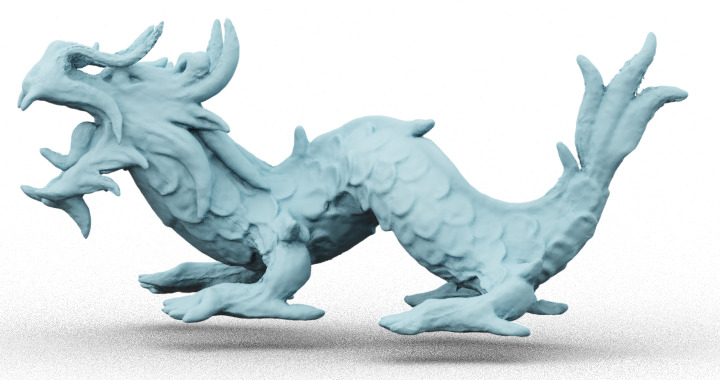} \\

        \includegraphics[width=0.46\columnwidth]{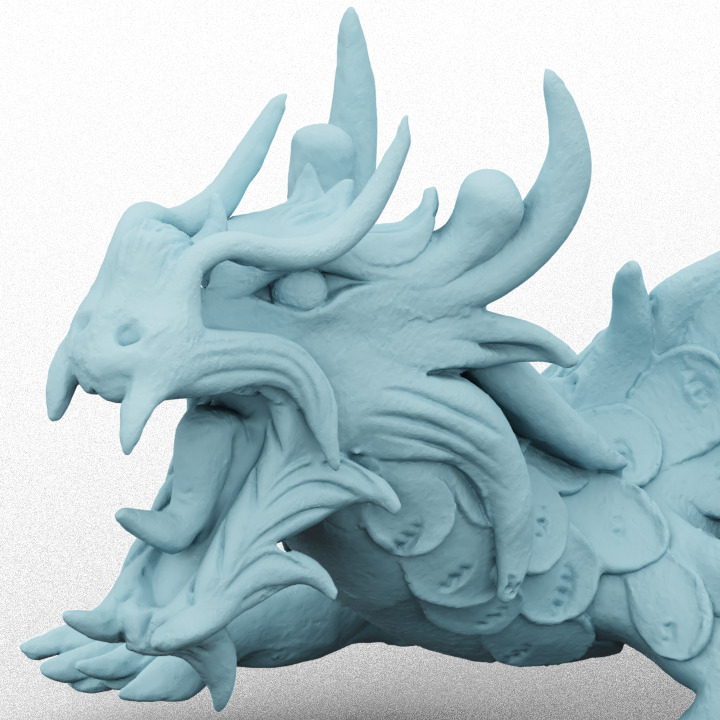} &
        \includegraphics[width=0.46\columnwidth]{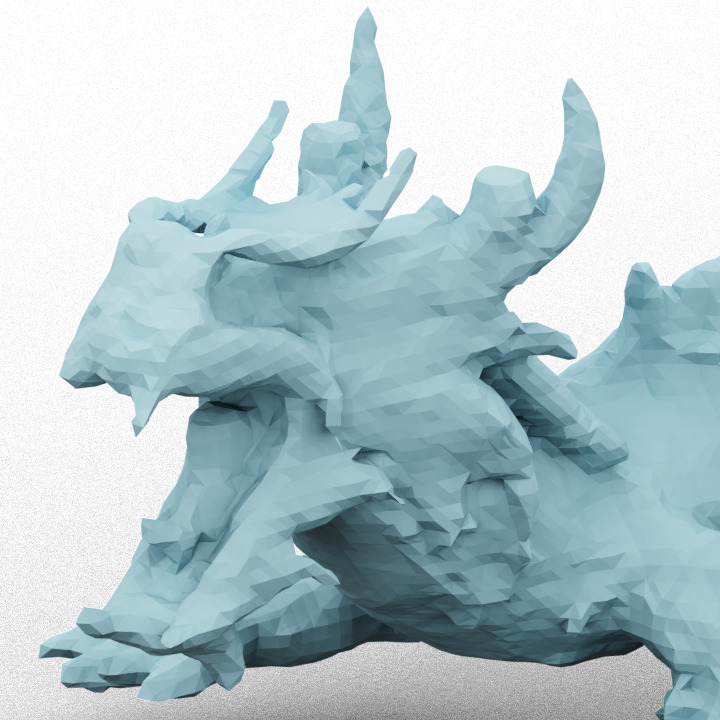} &
        \includegraphics[width=0.46\columnwidth]{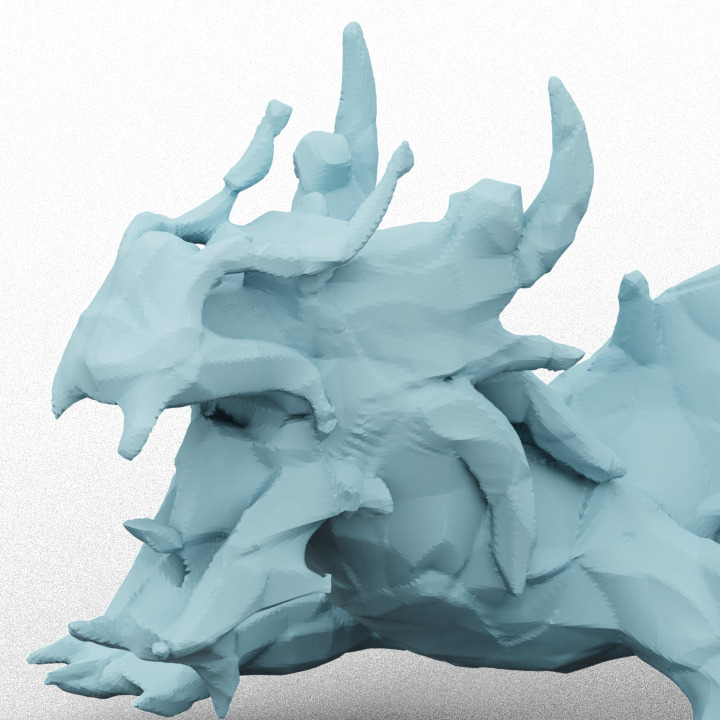} &
        \includegraphics[width=0.46\columnwidth]{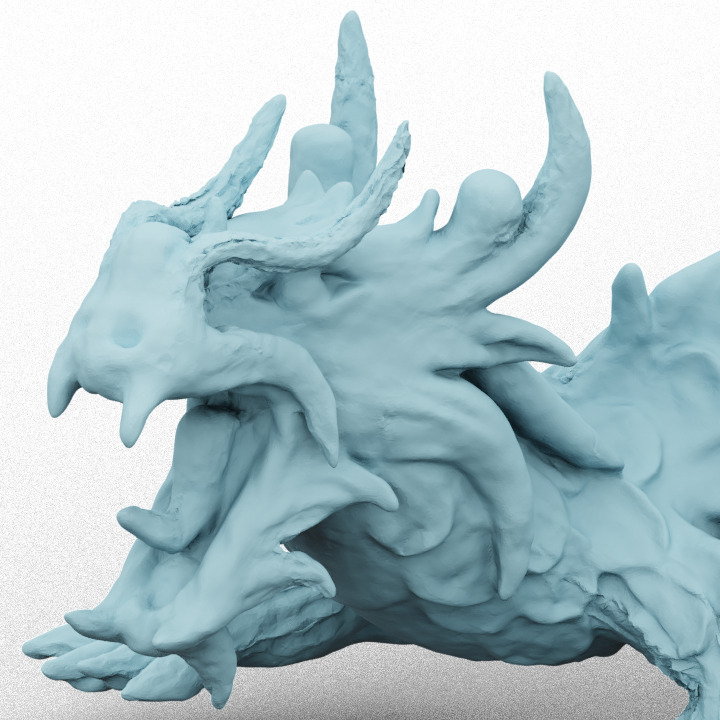} \\
        
        \includegraphics[width=0.46\columnwidth]{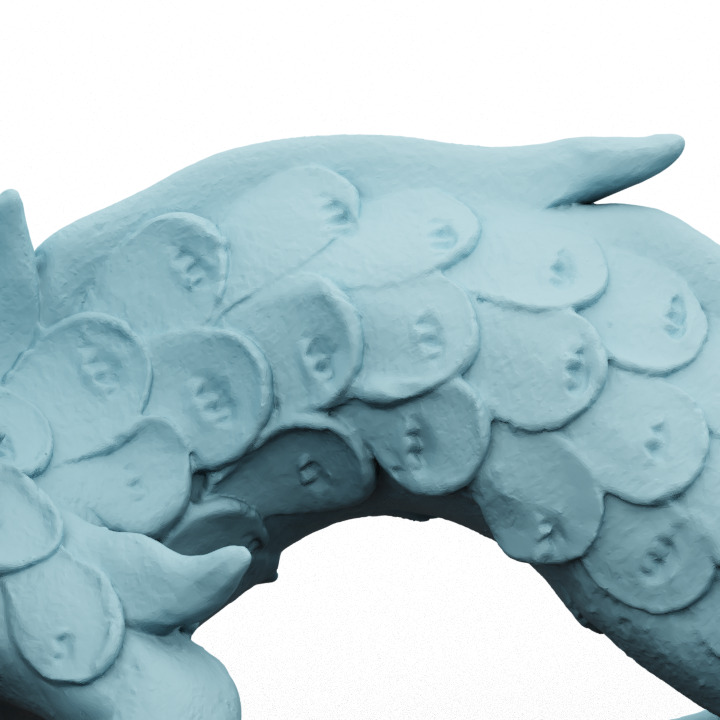} &
        \includegraphics[width=0.46\columnwidth]{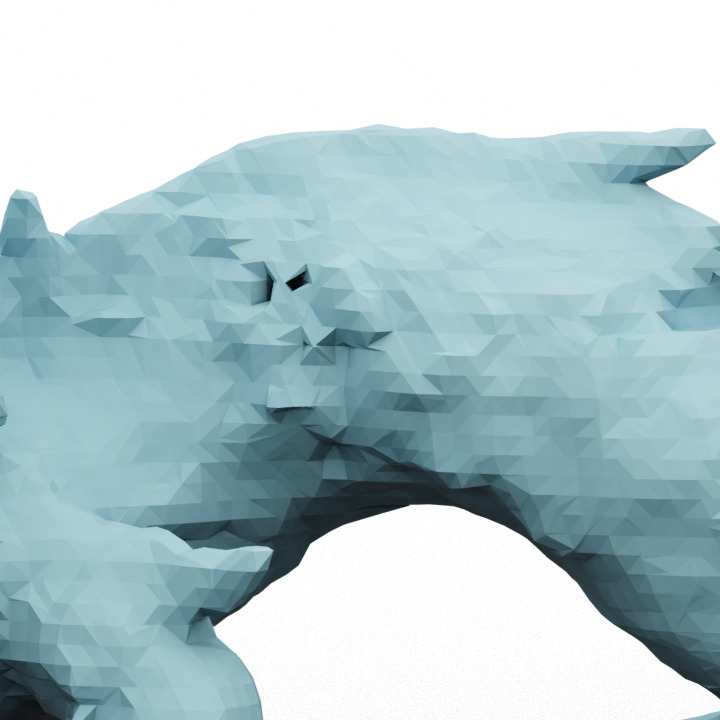} &
        \includegraphics[width=0.46\columnwidth]{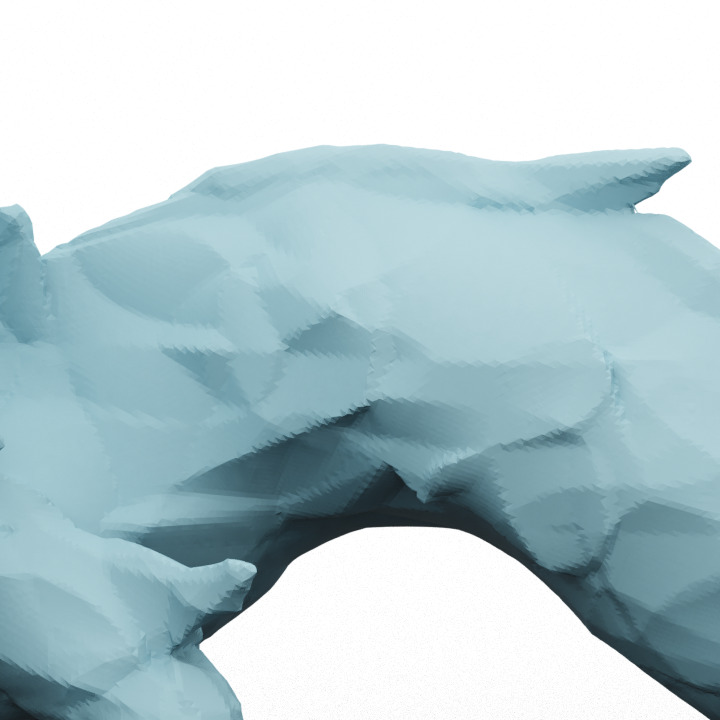} &
        \includegraphics[width=0.46\columnwidth]{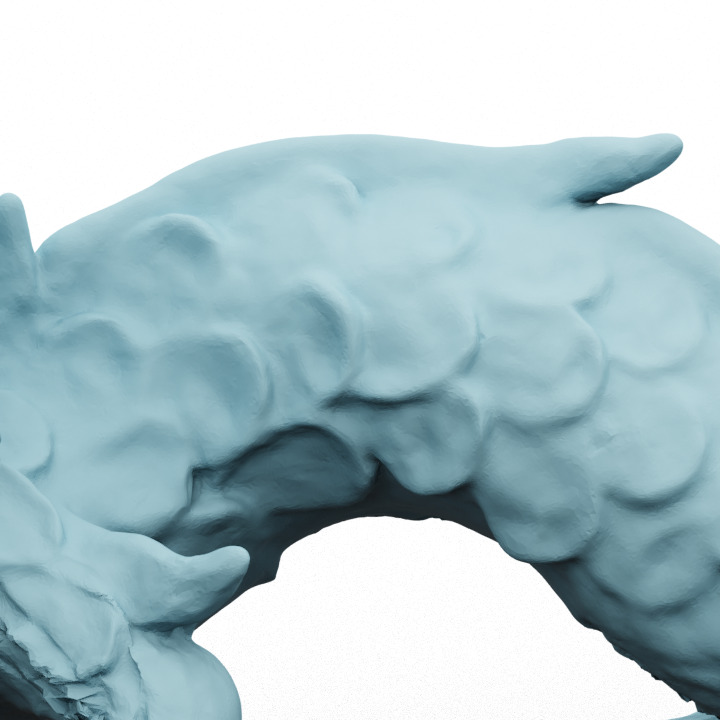} \\
        
        \textbf{(a)} Ground truth &
        \textbf{(b)} ACORN \cite{martel2021acorn} &
        \textbf{(c)} NGLOD \cite{takikawa2021neural} &
        \textbf{(d)} NCS (ours) \\
    \end{tabular}
    
    \caption{\textbf{Surface representation.} The reconstruction quality of our method, compared with ACORN~\cite{martel2021acorn} and NeuralLOD~\cite{takikawa2021neural} for two models, using the same number of network parameters on each method model size (100K parameters in this example). Our result exhibits higher accuracy and reconstruction of fine details, while not exhibiting artifacts such as artificial edges or aliasing.
    }
    \label{fig:reconstruction}
\end{figure*}

\section{Experiments}
We now detail the various experiments we performed. We compare our reconstruction quality to other methods, move on to ablations, and finish with a additional applications enabled by our representation.
\label{sec:experiments}

\begin{table}[t]
    \centering
    \caption{\label{tab:sparse_compare} Comparison of shape representations with 100K parameters wrt Chamfer distance $\downarrow$. Numbers are multiplied by $10^{3}$. }
    
    \setlength{\tabcolsep}{2.5pt}
    \begin{tabular}{r|c|c|c|c}
         & \cite{takikawa2021neural} & Sparse \cite{takikawa2021neural} & \cite{yifan2021geometryconsistent} & Ours \\
         \hline
         \hline
         Armadillo & 1.95 & 1.34 & 1.06 & \textbf{0.54} \\
         Bimba     & 2.30 & 2.07 & 2.09 & \textbf{1.04} \\
         Dino      & 1.70 & 1.55 & 2.55 & \textbf{1.48} \\
         Dragon    & 1.57 & 1.12 & 0.62 & \textbf{0.57} \\
         Grog      & 2.06 & 1.06 & \textbf{0.81} & 1.28 \\
         Seahorse  & 1.26 & 1.15 & - & \textbf{0.44} \\
         Elephant  & 4.06 & \textbf{2.24} & 3.93 & 2.49 \\
         Gargoyl   & 6.30 & - & 8.51 & \textbf{2.29} \\ 

    \end{tabular}
\end{table}

\paragraph{Comparison.} We compare our method against three state-of-the-art methods for neural shape representations:
(i)~ACORN~\cite{martel2021acorn}; (ii)~NGLOD~\cite{takikawa2021neural}, and (iii)~Neural Surface Maps (NSM)~\cite{morreale2021neural}. In the supplemental, we provide additional comparisons. We compare the methods on a variety of shapes with different amount and type of geometric details (see  Table \ref{tab:model_size}). In the main paper we provide comparisons on $5$ different shapes, with additional shapes in the supplementary material. Note that all models are scaled to a unit sphere. We evaluate performance along two main axes: 
(i)~The representation \emph{accuracy} is measured by the Bidirectional Chamfer distance, which computes the distance between output and ground truth surfaces. Lower values indicate more accurate representations. (ii)~The \emph{memory cost} is measured by the number of parameters required by a representation. Typically parameters are represented as 32 bit floats, so multiplying by $4$ gives the number of bytes.

As can be seen in Figure \ref{fig:graph_reconstruction}, for the same number of parameters, our method achieves significantly higher accuracy than the state of the art. Furthermore, in all cases but one (Lucy), our method's accuracy exceeds all others methods even when using $10$ times less parameters.
As shown in Figure \ref{fig:reconstruction}, our method preserves detail such as the dragon's scales and Bimba's braids much more accurately than both ACORN and NGLOD.
Furthermore, both competing methods exhibit discretization artifacts or noise, while our method provides a smooth, artifact-free surface.

We offer additional comparisons against the concurrent method IDF~\cite{yifan2021geometryconsistent} in Table~\ref{tab:sparse_compare}. Further visual comparison is offered in the supplementary material.
IDF~\cite{yifan2021geometryconsistent} fails to correctly describe the iso-surface of Dino and Elephant. Note that iso-surfaces produced by IDF include excess geometry that wraps around the shape. To offer a numerical comparison, we manually removed the additional excess surface. IDF completely fails to represent the Seahorse and hence we leave a `-' in the table.

\begin{figure}
    \centering
    \setlength\tabcolsep{1.5pt} 
    \begin{tabular}{cccc}
        \includegraphics[width=0.21\columnwidth]{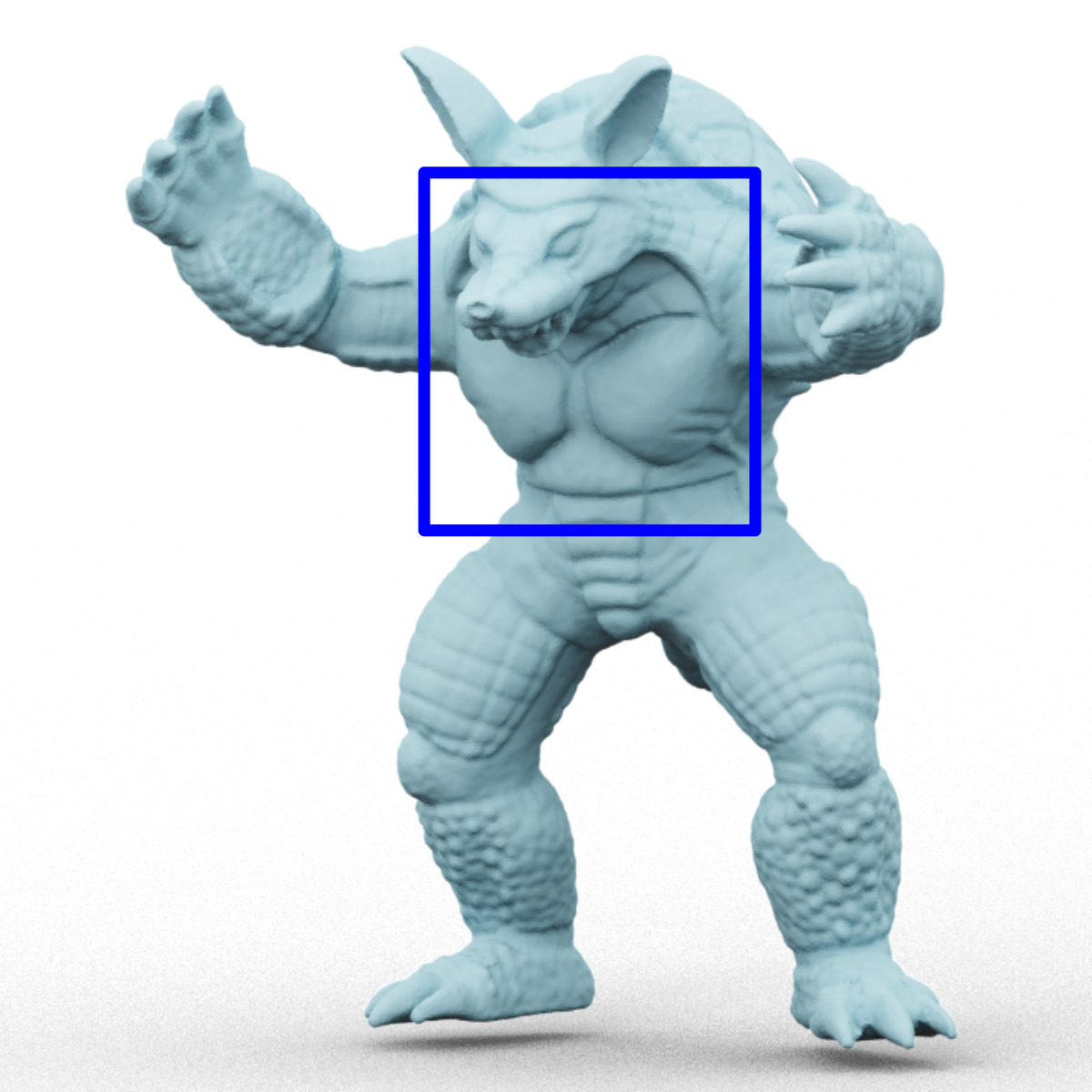} &
        \includegraphics[width=0.23\columnwidth]{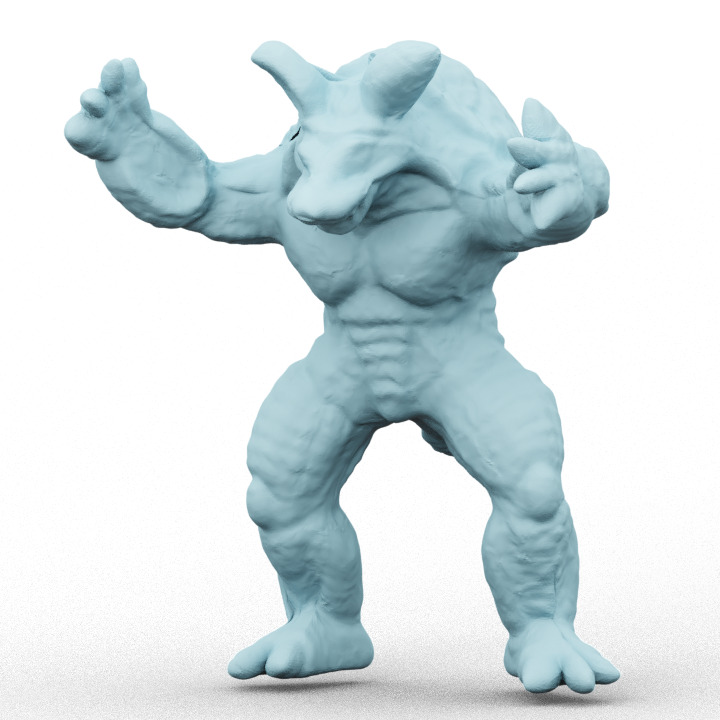} & 
        \includegraphics[width=0.23\columnwidth]{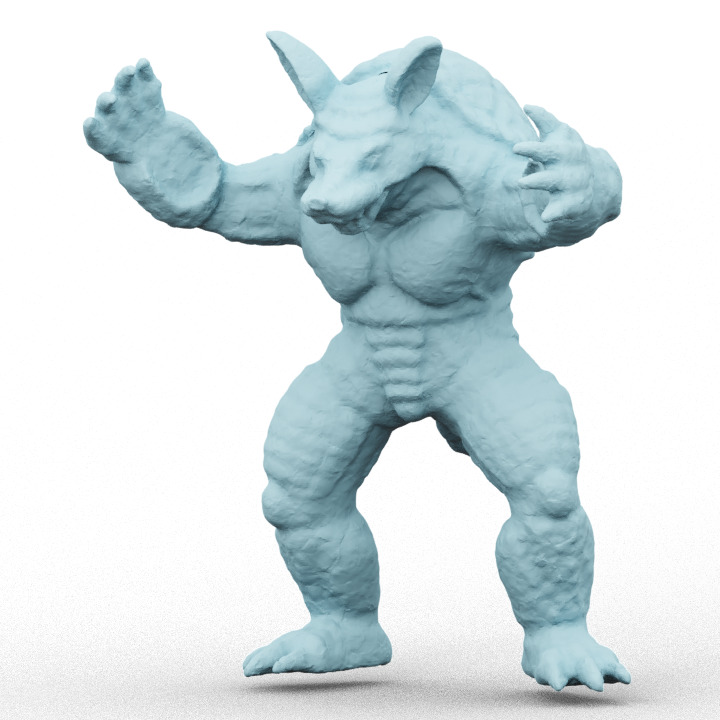} & 
        \includegraphics[width=0.23\columnwidth]{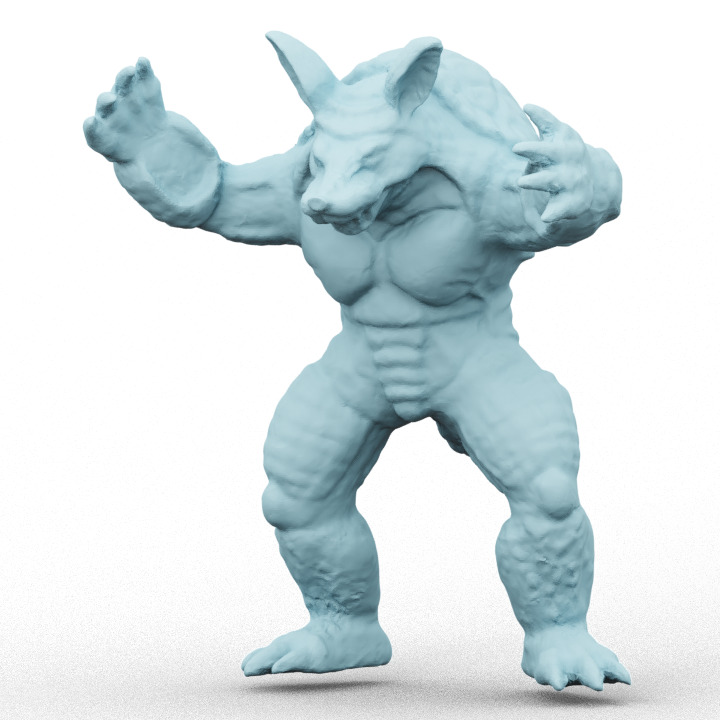} \\
        \includegraphics[width=0.23\columnwidth]{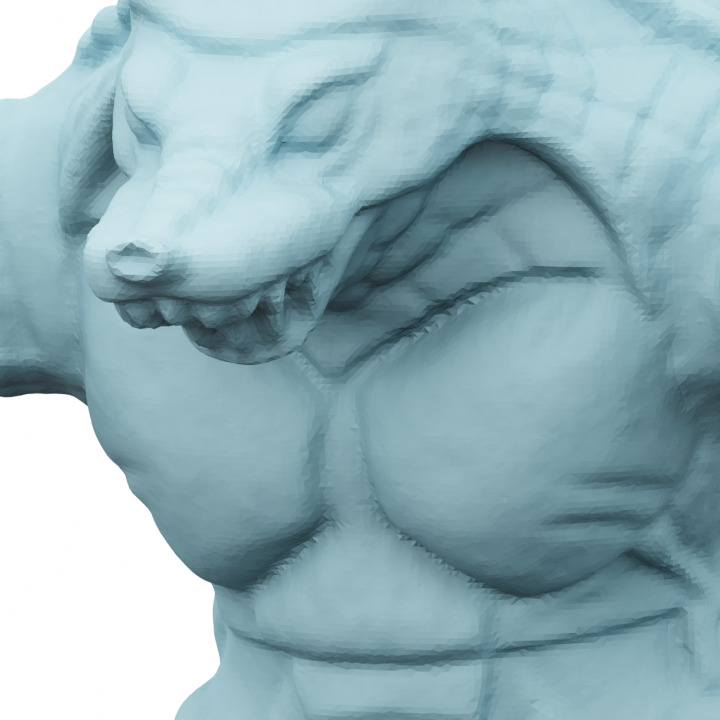} &
        \includegraphics[width=0.23\columnwidth]{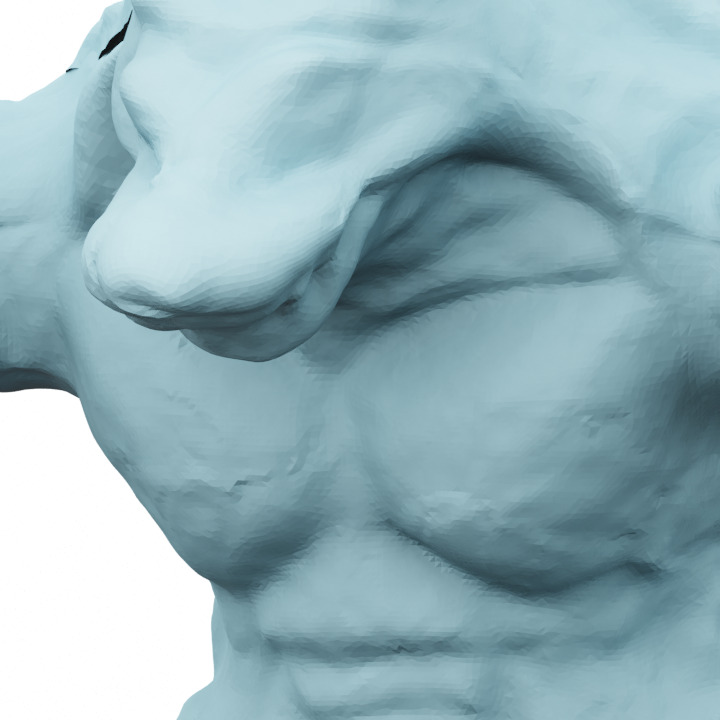} & 
        \includegraphics[width=0.23\columnwidth]{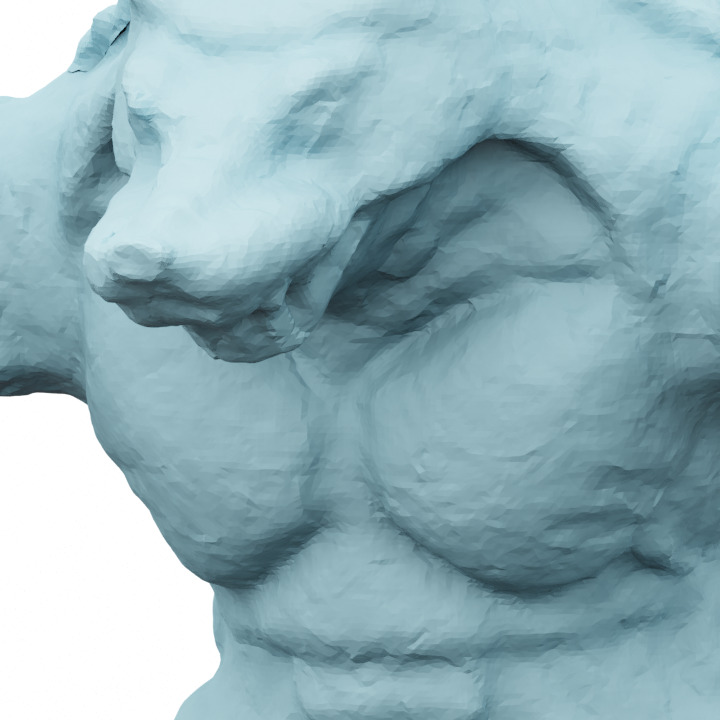} & 
        \includegraphics[width=0.23\columnwidth]{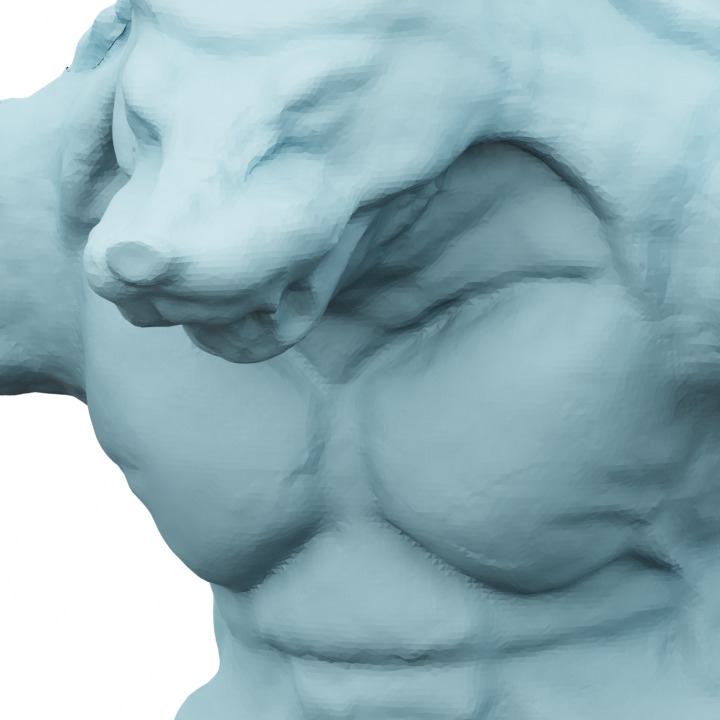} \\
        \textbf{(a)} GT &
        \textbf{(b)} normal &
        \textbf{(c)} PCA &
        \textbf{(d)} Ours \\
    \end{tabular}
    
    \caption{Ablation study.
    (a)~the ground truth model; (b)~adding scalar displacement along the normal direction (to a learned base coarse model) yields smoothed-out results; (c)~adding displacement vectors to a per-patch canonical coordinate frame (established using patch's PCA axes) yields artifacts and surface ripples; (d)~our reconstruction is sharp and does not exhibit artifacts.  
    }
    \label{fig:pca_compare}
\end{figure}

\paragraph{Ablation.} Figure~\ref{fig:pca_compare} evaluates the necessity of various components in our framework: (i)~\textit{scalar displacements}: we restrict the fine network to only apply scalar displacements along normal directions, instead of displacement vectors as used in the full model. This significantly hinders the ability of the fine model to add details on top of the coarse model, leading to an oversmoothed result that resembles the coarse model; (ii)~\textit{PCA only}: we remove the coarse model and use only the fine model. For the local reference frame $\mathbf{F}$ required by the fine model, we pre-compute and store the PCA frame of the ground truth patch.
This causes the fine model to spend its capacity on re-creating the coarse geometry, and as a result, artifacts and ripples appear in the reconstruction.

\pgfplotstableread[col sep=space]{armadillo.txt}\armadillodata


\pgfplotstableread[col sep=space]{bimba.txt}\bimbadata


\pgfplotstableread[col sep=space]{dino.txt}\dinodata


\pgfplotstableread[col sep=space]{lucy.txt}\lucydata


\pgfplotstableread[col sep=space]{dragon.txt}\dragondata


\pgfplotstableread[col sep=space]{ours.txt}\oursdata

\begin{figure}[t]
\centering
\setlength\tabcolsep{1.5pt} 
\begin{tabular}{cc}
    \begin{tikzpicture}[scale=0.48]
    \node at (2.5,5.2) {Armadillo};
    \begin{axis}[
      xmode=log,log basis x=10,
      ymode=log,log basis x=10,
      xlabel=\# parameters,
      ylabel=Chamfer distance,
      xtick=data,
      xticklabels from table={armadillo.txt}{params}
      ]
    \addplot table [x=step,y=acorn]{\armadillodata};
    \addlegendentry{ACORN}
    \addplot table [x=step,y=nglod]{\armadillodata};
    \addlegendentry{NGLOD}
    
    \addplot[mark=*] coordinates {(1,0.001411)};
    \addlegendentry{NSM}
    
    \addplot table [x=step,y=armadillo]{\oursdata};
    \addlegendentry{Ours}
    
    \end{axis}
    \end{tikzpicture}
&
    \begin{tikzpicture}[scale=0.48]
    \node at (2.5,5.2) {Bimba};
    \begin{axis}[
      xmode=log,
      ymode=log,
      xlabel=\# parameters,
      xtick=data,
      xticklabels from table={bimba.txt}{params}
      ]
    
    \addplot table [x=step,y=acorn]{\bimbadata};
    \addlegendentry{ACORN}
    \addplot table [x=step,y=nglod]{\bimbadata};
    \addlegendentry{NGLOD}
    \addplot[mark=*] coordinates {(1,0.00161)};
    \addlegendentry{NSM}
    
    \addplot table [x=step,y=bimba]{\oursdata};
    \addlegendentry{Ours}
    \end{axis}
    \end{tikzpicture}
\\

    \begin{tikzpicture}[scale=0.48]
    \node at (2.5,5.2) {Ankylos.};
    \begin{axis}[
      xmode=log,
      ymode=log,
      xlabel=\# parameters,
      ylabel=Chamfer distance,
      xtick=data,
      xticklabels from table={dino.txt}{params}
      ]
    
    \addplot table [x=step,y=acorn]{\dinodata};
    \addlegendentry{ACORN}
    \addplot table [x=step,y=nglod]{\dinodata};
    \addlegendentry{NGLOD}
    
    \addplot[mark=*] coordinates {(1,0.001478)};
    \addlegendentry{NSM}
    
    \addplot table [x=step,y=dino]{\oursdata};
    \addlegendentry{Ours}
    \end{axis}
    \end{tikzpicture}
&
    \begin{tikzpicture}[scale=0.48]
    \node at (2.5,5.2) {Dragon};
    \begin{axis}[
      xmode=log,
      ymode=log,
      xlabel=\# parameters,
      xtick=data,
      xticklabels from table={dragon.txt}{params}
      ]
    
    \addplot table [x=step,y=acorn]{\dragondata};
    \addlegendentry{ACORN}
    \addplot table [x=step,y=nglod]{\dragondata};
    \addlegendentry{NGLOD}
    
    \addplot[mark=*] coordinates {(1,0.000641)};
    \addlegendentry{NSM}
    
    \addplot table [x=step,y=dragon]{\oursdata};
    \addlegendentry{Ours}

    \end{axis}
    \end{tikzpicture}
\\

\end{tabular}

\caption{\label{fig:graph_reconstruction} 
Reconstruction quality versus Model complexity for different models. Note that our method achieves better reconstruction quality with significantly lower memory footprint. 
Note that values are reported in logscale. Bottom row shows our results. Given our reconstructions, we do not use more than 1M parameters. 
}

\end{figure}
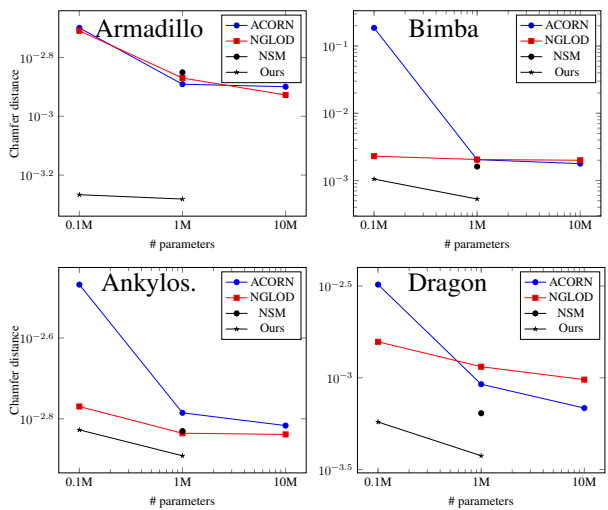

\begin{table}[t]
    \centering
    \caption{\label{tab:model_size}Evaluation and distribution of the network parameters between the different modules in our architecture, for various models tested in the paper, of size 100K. The latent codes use the majority of parameters, while the other components are self-contained.}
    \setlength{\tabcolsep}{2.5pt}
    \small
    \begin{tabular}{r|c|c|c|c|c|c|c}
        &&& Coarse & \multicolumn{3}{c|}{Fine} & TOT. \\
        &\#V&\#F& MLP & Code & CNN & MLP & params \\
        \hline
        \hline
        Armadillo & 172K & 346K & 13K &  93K & 6K & 467 & 113K \\
        Bimba     & 50K & 100K &  9K & 115K & 6K & 467 & 130K \\
        Lucy      & 877K & 1753K & 13K &  96K & 4K & 435 & 114K \\
        Dino  & 26K & 51K &  4K & 120K & 6K & 467 & 130K \\
         Dragon   & 451K & 902K & 13K &  99K & 6K & 467 & 119K \\
    \end{tabular}
\end{table}

\begin{figure}[t]
    \centering
    \begin{tabular}{ccc}
    \includegraphics[width=0.47\columnwidth]{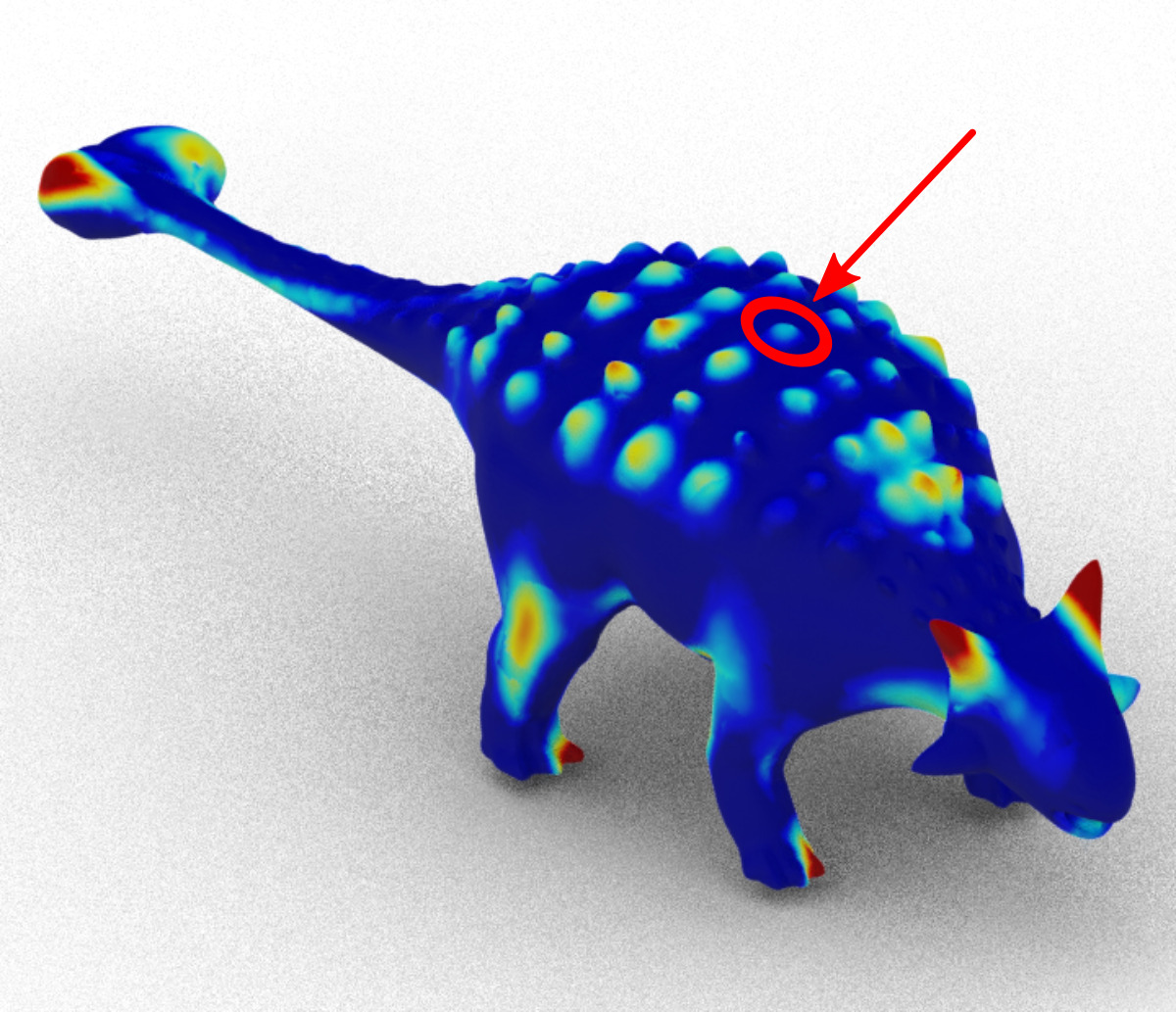} &
    \includegraphics[width=0.47\columnwidth]{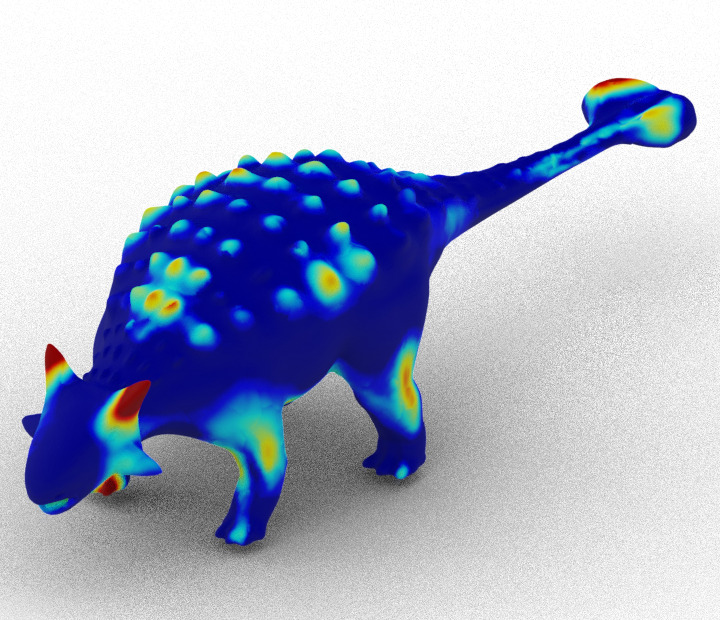} \\
    \end{tabular}
    \caption{
    Our method yields interpretable convolutional kernels:
    we select one spike (highlighted) on the dino, identify the CNN features that are strongly active in its region, and then identify other regions where the same features are active. High correlation (hotter colors) implies regions with similar geometric details.
    }
    \label{fig:interp}
\end{figure}

\begin{figure}[t]
    \centering
    \setlength\tabcolsep{1.5pt} 
    \begin{tabular}{ccc}
        \includegraphics[width=0.28\columnwidth]{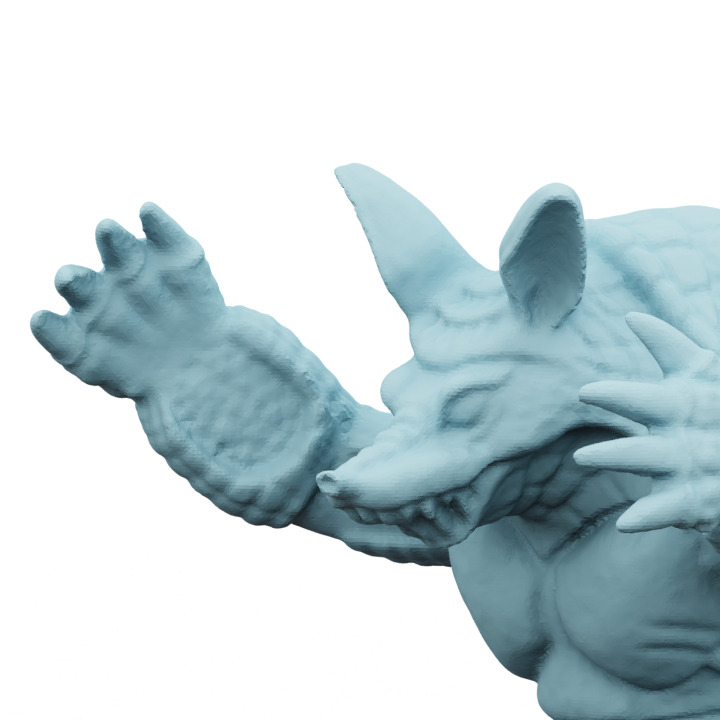} &
        \includegraphics[width=0.28\columnwidth]{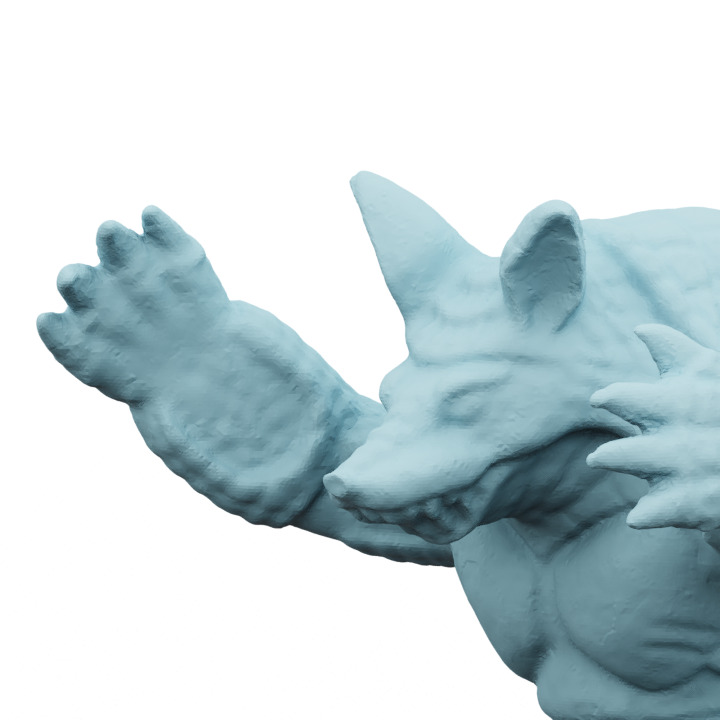} &
        \includegraphics[width=0.28\columnwidth]{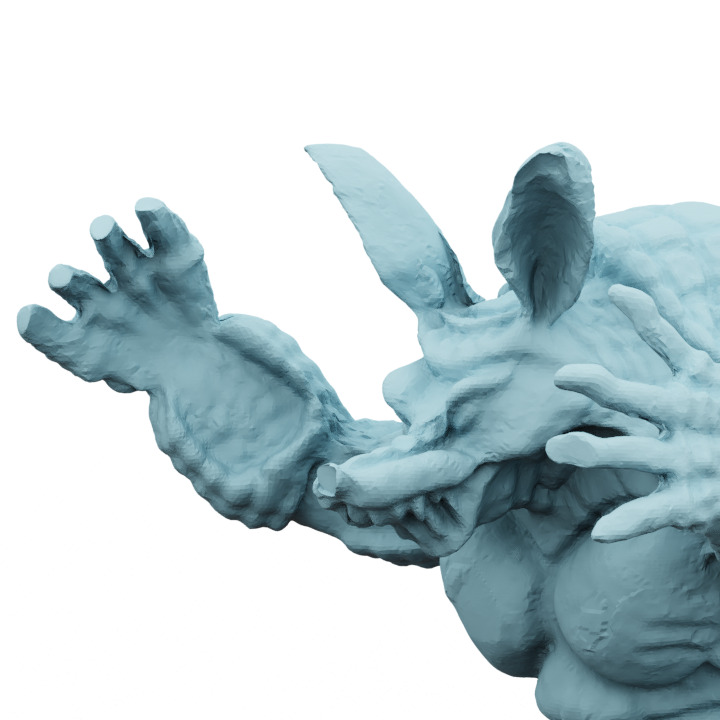} \\
        \includegraphics[width=0.28\columnwidth]{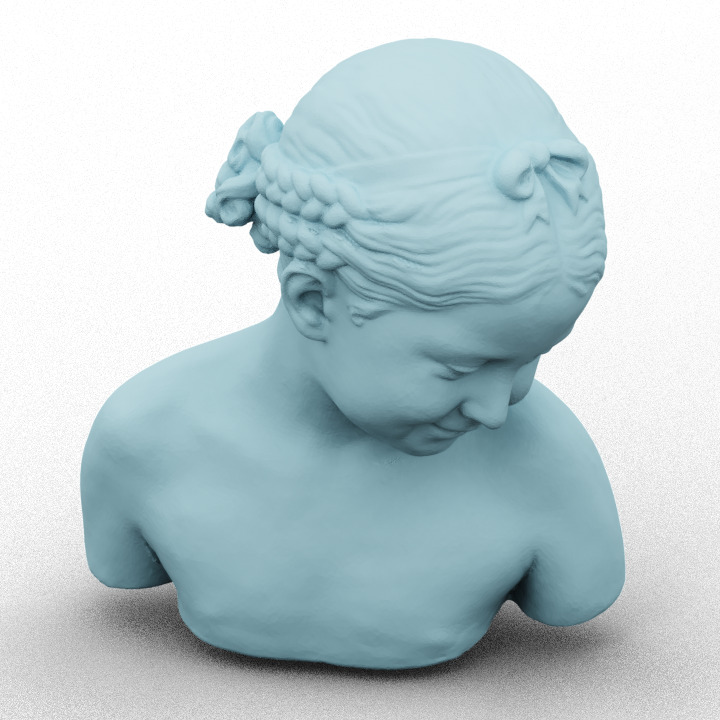} &
        \includegraphics[width=0.28\columnwidth]{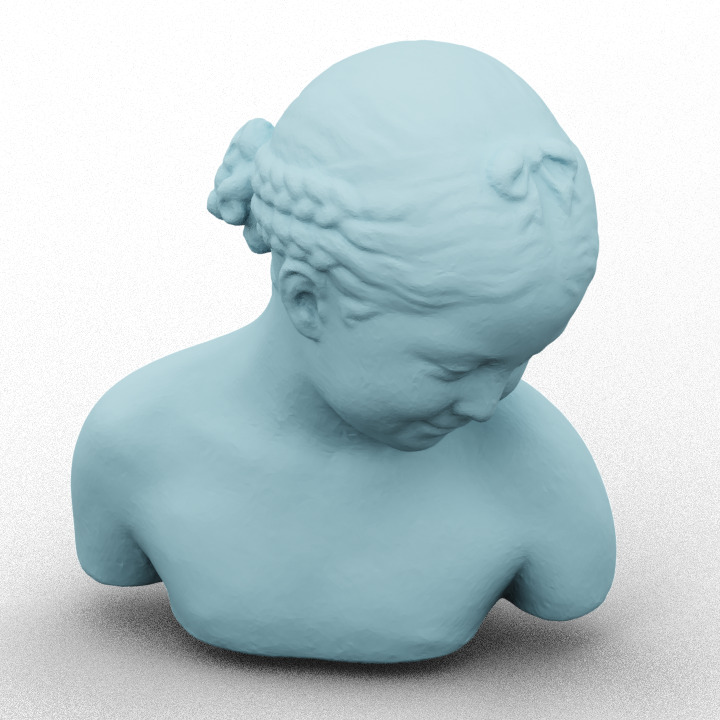} &
        \includegraphics[width=0.28\columnwidth]{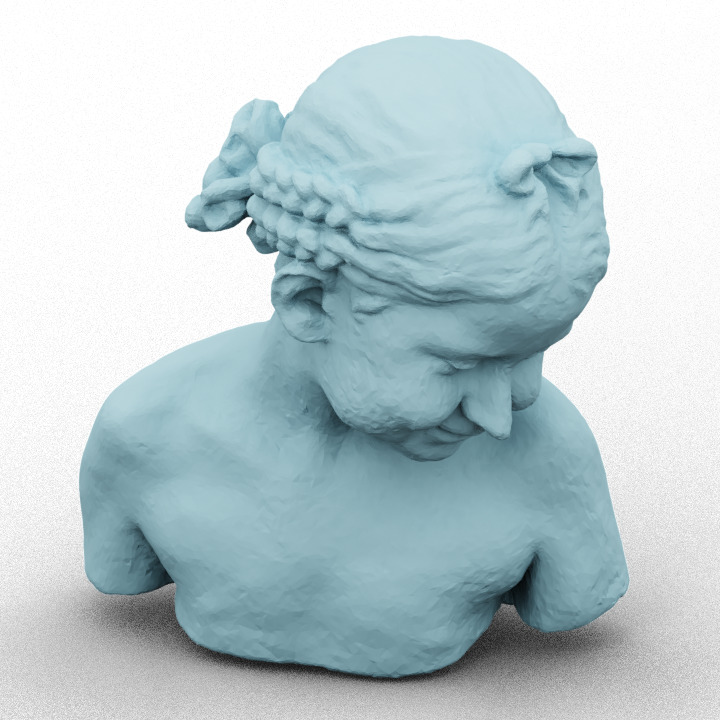} \\
        \includegraphics[width=0.28\columnwidth]{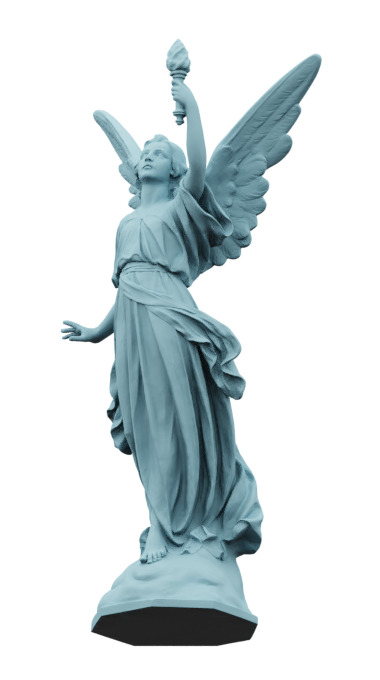} &
        \includegraphics[width=0.28\columnwidth]{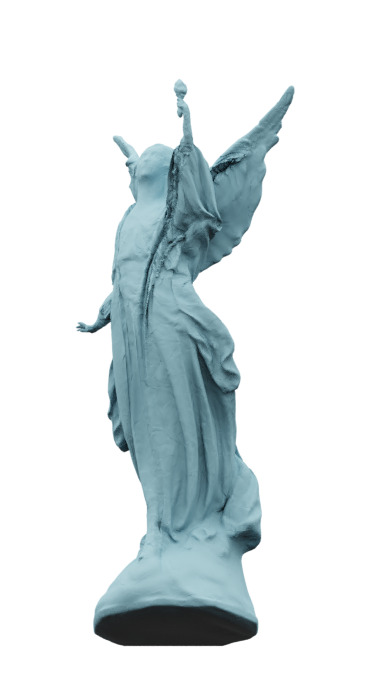} &
        \includegraphics[width=0.28\columnwidth]{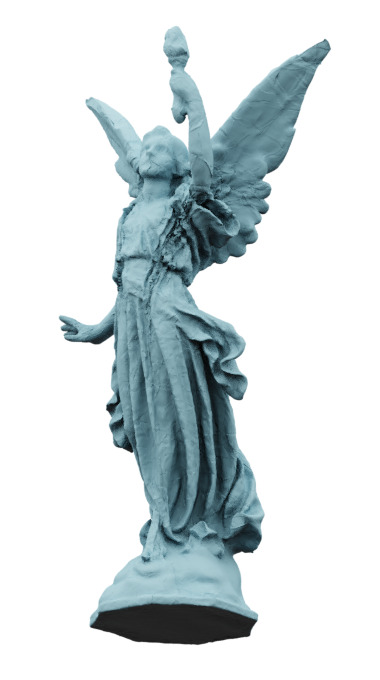} \\
       
        GT & Smoothing & Sharpening \\
    \end{tabular}
    
    \caption{\textbf{Sharpening and smoothing.} 
    Our NCS naturally decomposes shapes into coarse shapes and fine details. Boosting or suppressing the fine details and reconstructing the shapes, naturally results in exaggeration or smoothing of surface features.  
    }
    \label{fig:sharp_smooth}
\end{figure}

\paragraph{Feature enhancement.} \label{ss:exp_features}

Since the CNN encodes local geometric detail, we can edit geometric detail by manipulating the CNN's feature maps.
In Figure \ref{fig:sharp_smooth} we perform feature enhancement on a few models by scaling up the CNN output, before feeding it into the MLP of the fine model. Similarly, we can perform smoothing, by scaling down the same features.

\begin{figure}[t]
    \centering
    \setlength\tabcolsep{1.5pt} 
    \begin{tabular}{cc}
        \includegraphics[width=0.4\columnwidth]{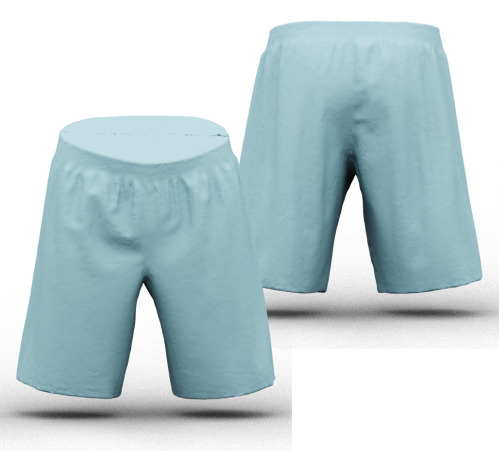} &
        \includegraphics[width=0.4\columnwidth]{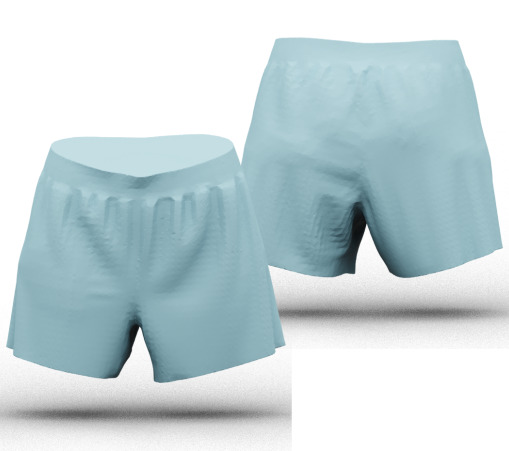} \\
        \textbf{(a)} Source & \textbf{(b)} Target \\
        \includegraphics[width=0.4\columnwidth]{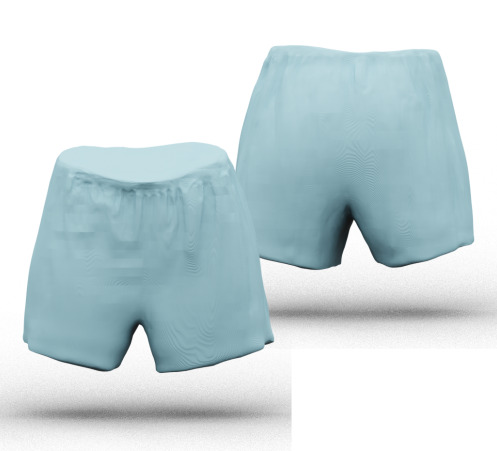} &
        \includegraphics[width=0.4\columnwidth]{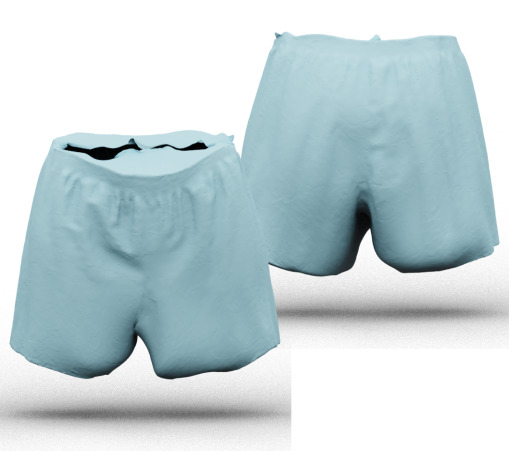} \\
        \textbf{(c)} IDF~\cite{yifan2021geometryconsistent} &
        \textbf{(d)} Ours \\
    \end{tabular}
    
    \caption{\textbf{Detail transfer.} 
    Our architecture intrinsically decomposes shapes into coarse base models and associated geometric details, which allows us to transfer learned details from one model to another base shape. In this case, from one pair of pants (a), to another pair of pants. The fitter coarse model for each of the two pairs is shown at the top. Here we compare our detail transfer results with those of a concurrent work~\cite{yifan2021geometryconsistent}. To transfer details we replace the source coarse model with the target coarse one, and reconstruct the shape. Note, this is possible because the global geometry images, source and target, are aligned. In case of misalignment, an inter-surface map between the coarse models could be computed using, e.g., \cite{morreale2021neural}. 
    }
    \label{fig:transfer}
\end{figure}

\paragraph{Detail transfer.} \label{ss:exp_transfer}

Our disentanglement of coarse and fine detail enables us to perform detail transfer, similarly to IDF\cite{yifan2021geometryconsistent}. Figure \ref{fig:transfer} show results of our method, transferring creases from one model to the other. Similarly to IDF, we achieve this by training one coarse network for the source shape and one for the target shape. The fine network is trained to accurately fit the source shape. We then transfer the details to the target shape with a forward pass using the source global and local parameterization.

\paragraph{Interpretability of the kernels.}
The use of a CNN in the fine branch of our network's architecture leads to interpretable kernels, i.e., specific kernels react to specific details of the geometry. In Figure \ref{fig:interp} we show an example in which we select a region on the model, find features that are activated strongly in that region, and then highlight other areas in which those features are activated. As we can see, the features associated with one of the Dino's spikes also affect all other spikes. This shows that our kernels are reused across the model,
explaining our network's ability to represent detailed models with a smaller number of network parameters than previous methods. This may also lead to future work where we use the same kernels to a larger collection of surface, to learn more specific and robust features. 

\paragraph{Implementation details.}

Patches are found by randomly sampling patch centers $c_i$ on $S$ and selecting all points within a geodesic radius $\rho$: $R_i = \{p\ |\ d^\text{geo}(p, c_i) le \rho\}$. We use an iterative approach: after creating a patch, all points inside the patch are marked as forbidden for the following patch centers with a probability $\eta$, which controls the amount of overlap between patches. In our experiments, we set $\eta = 0.5$ and $\rho=0.04$ times the maximum extent of $S$ along any coordinate axis. Figure~\ref{fig:effectOfPatchSize} shows the effect of the choice of number of patches.

In terms of training performance, we observed that we can achieve better results with a training schedule that starts by warming up the coarse model before slowly ramping up training of the fine model:
\begin{gather}
\mathcal{L} = (1-\lambda) \mathcal{L}_\text{joint} + \lambda \mathcal{L}_\text{reg} \\
\text{with } \mathcal{L}_\text{reg} = \int_{Q_S} \| g^c_\phi(q) - s(q) \|_2^2 \ dq. \nonumber
\end{gather}
We start with $\lambda=1$ and progressively decrease $\lambda \rightarrow 0$ over the warm-up phase, which lasts for $100$K iterations. At the same time, we increase the learning rate of the fine model and decrease the learning rate of the coarse model over the warm-up phase: the learning rate of the coarse model follows a cosine annealing schedule~\cite{loshchilov2016sgdr}, down to a minimum learning rate of $0$ at the end of the warm-up phase, while the learning rate of the fine model is set to $1e-4$ minus the coarse learning rate. The total number of iteration varies based on the complexity of the model, \eg, between 800K to 1.4M iterations, using the RMSProp optimizer~\cite{graves2013generating}. 

\section{Conclusions}

Neural convolutional surfaces enable faithfully representing a given surface via a neural network, with higher accuracy and a smaller network capacity (e.g., $10$-$80$x) compared to multiple state-of-the-art alternatives. Key to our method is an inductive bias in the network architecture that results in a split representation, with an MLP producing a coarse abstraction of the shape,  and a fine detail CNN-like layer that adds geometric displacements based on the local reference frame the local UV charts. 

We demonstrate that this coarse-fine disentanglement emerges naturally, without any intermediate supervision, and leads to the fine module reusing its convolutional kernels, which in turn enable meaningful geometric operations like mesh smoothing and feature exaggeration. 

\paragraph{Limitations and Future Work.} 
While the CNN based architecture leads to significant compression by reusing the kernels across object-centric local coordinate frames, the kernels themselves are 
still regular, 2D Euclidean image kernels, and hence are not rotationally invariant, as they ideally should be to handle geometry. This hinders perfect reuse of kernels across the shape, e.g., in cases of asymmetric features that are reoriented on the shape (rotated on the local tangent space), for example, the scales on the dragon. Furthermore, the kernels cannot be reused to capture local \emph{deformations} of the underlying geometric details. Lastly, we note that in some cases our pipeline, in absence of intermediate supervision, may associate coarse structures as fine, e.g., Lucy's (the angel) torch in Figure \ref{fig:sharp_smooth} is reconstructed mainly by the fine module of our networks, and as a result is reduced in size when the details are smoothed. 

While we focused on faithfully representing individual shapes for the scope of this work, we intend to followup the next goal, of capturing \emph{distributions} of shapes. We observe that geometric details is often reused across shapes and hence we can aspire to learn a universal dictionary of CNN detail kernels, that can then be applied across a diverse set of shapes, where the global structures are captured by shape-specific coarse abstractions. Such a universal dictionary of local geometric details will be a close analog of low level features learnt on images (e.g., VGG features learnt using ImageNet), which, in turn, will enable both manipulation or transfer of details, as well as compression of shapes with a fixed universal shape-dictionary.

\paragraph*{Acknowledgements.}
LM was partially supported by the UCL Centre for AI and the UCL Adobe PhD program. This project has received funding from the European Union’s Horizon 2020 research and innovation programme under the Marie Skłodowska-Curie grant agreement No 956585.

{\small
\bibliographystyle{ieee_fullname}
\bibliography{main}
}

\cleardoublepage

\beginsupplement
\section{Comparisons}

\begin{figure}[t]
    \centering
    \setlength\tabcolsep{1.5pt} 
    \begin{tabular}{cccc}
         \includegraphics[width=0.25\columnwidth]{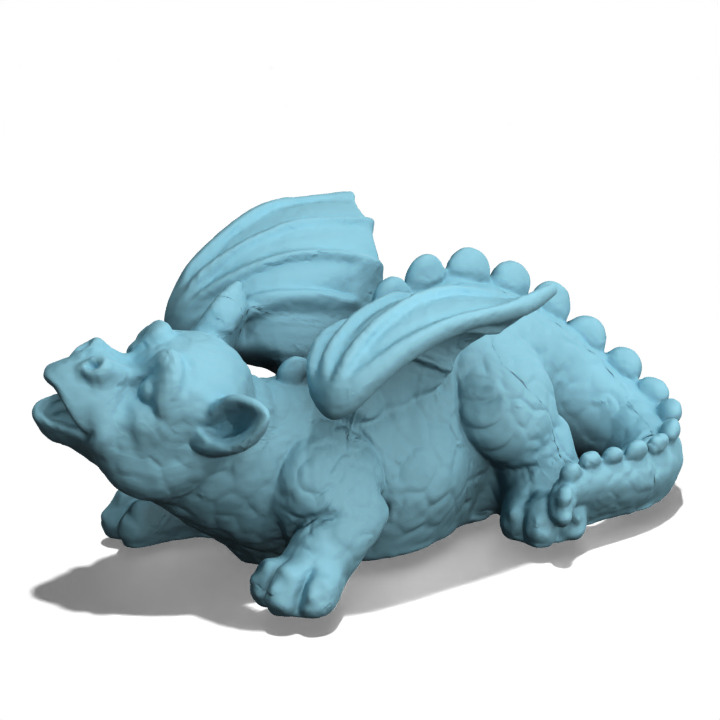} &
         \includegraphics[width=0.25\columnwidth]{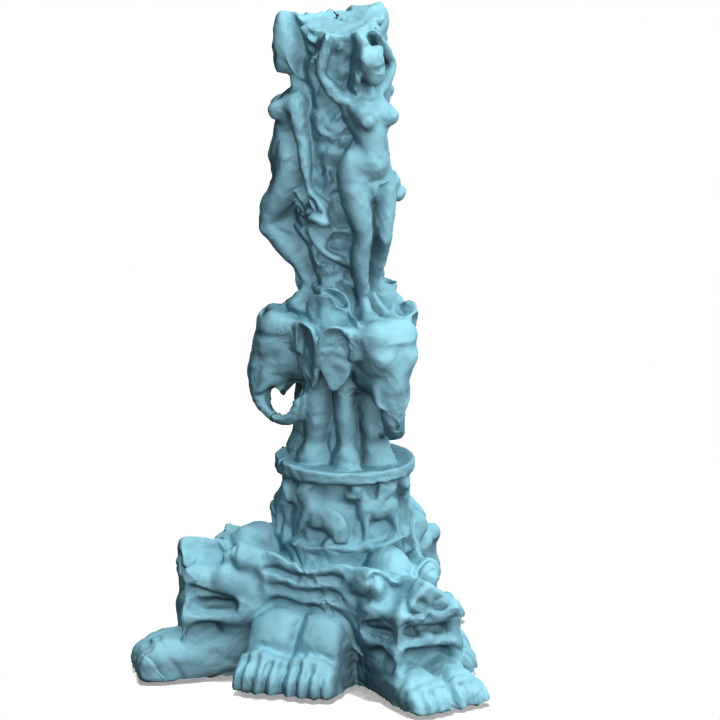} &
         \includegraphics[width=0.25\columnwidth]{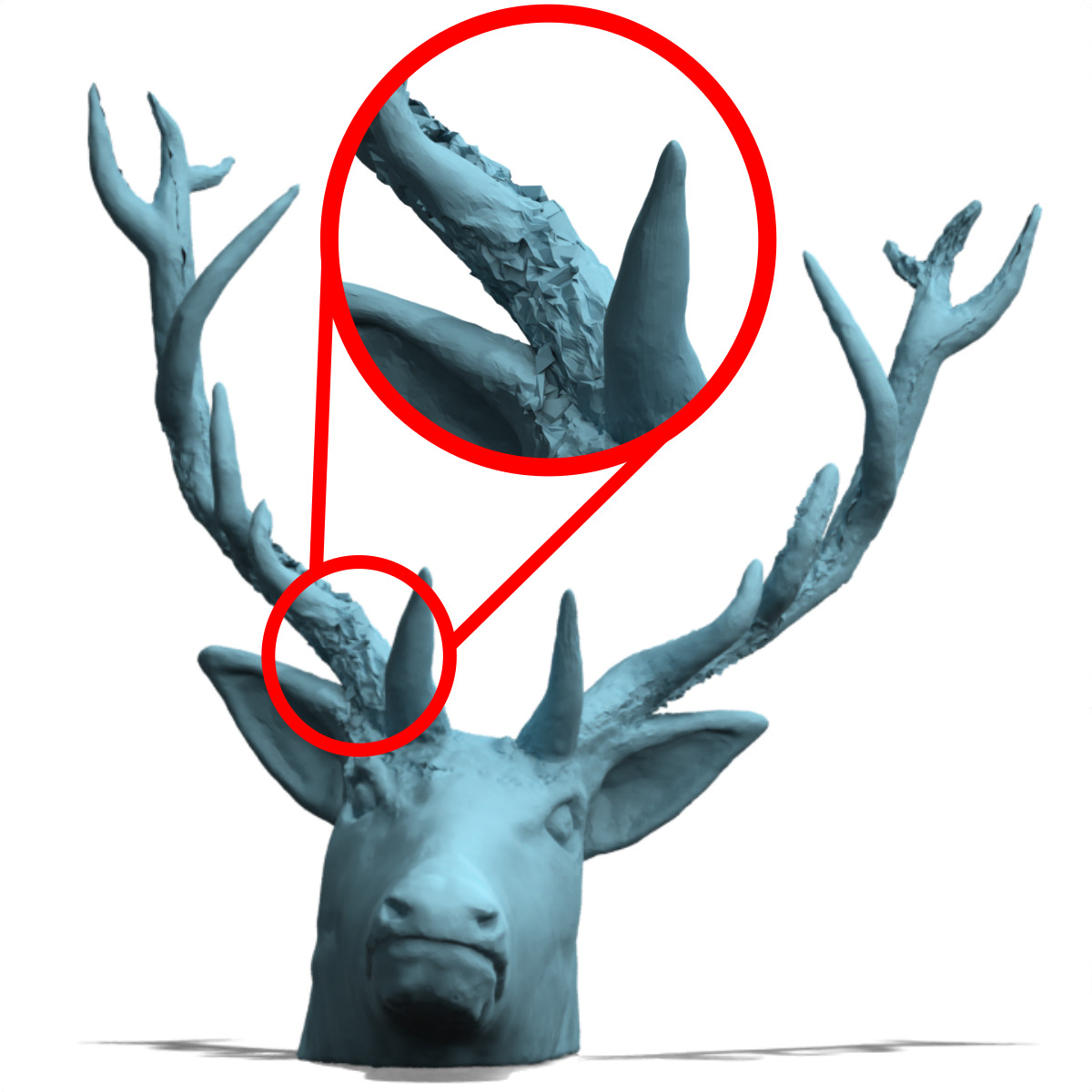} &
         \includegraphics[width=0.25\columnwidth]{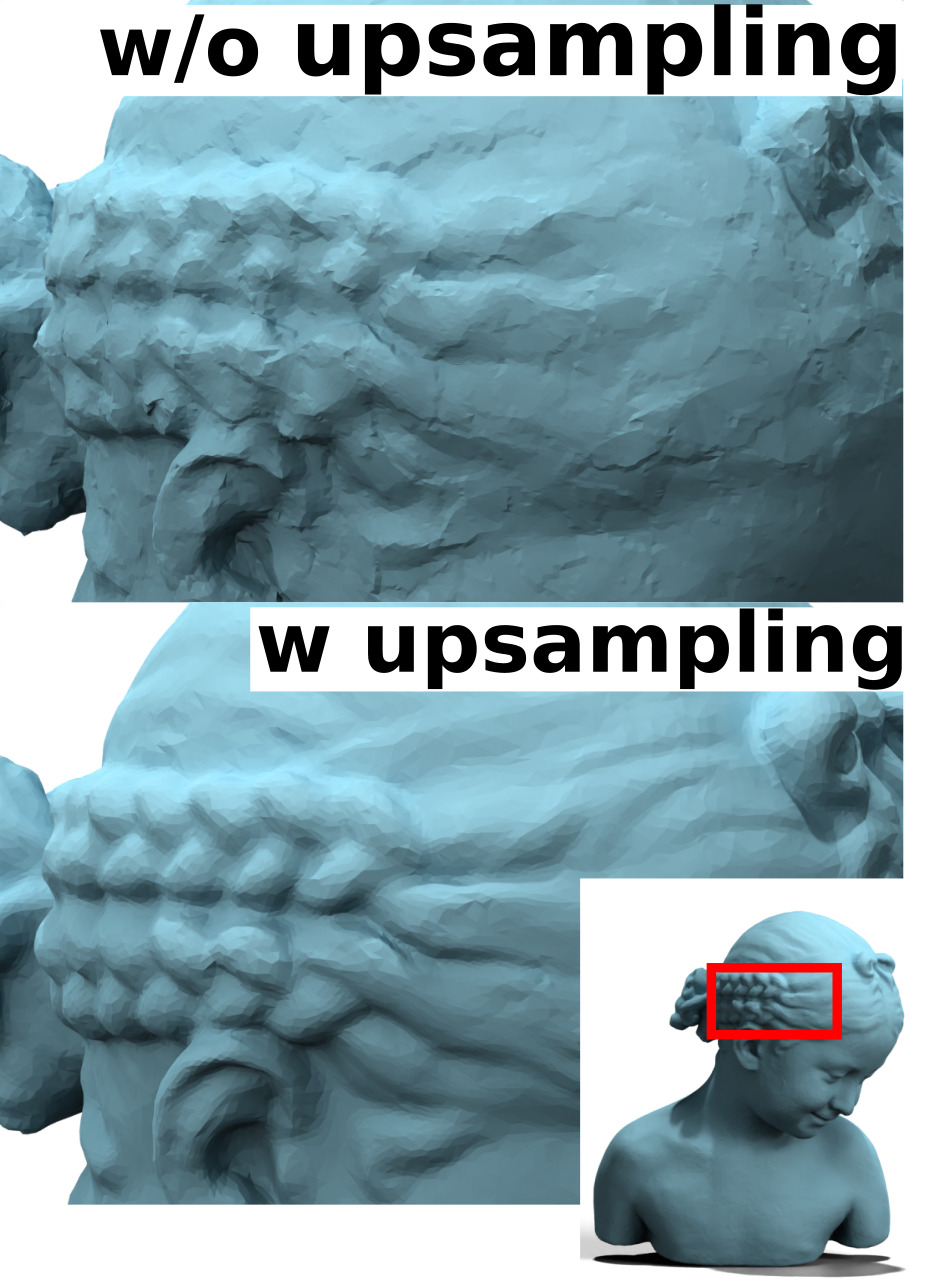} \\
         \textbf{(a)} &
         \textbf{(b)} &
         \textbf{(c)} &
         \textbf{(d)} \\
         Genus 2 &
         Genus 3 &
         ThinStructure &
         NoUpsampling \\
    \end{tabular}
    
    \caption{Results on high genus, thin structures, w/o upsampling.}
    \label{fig:topology}
\end{figure}

\paragraph{Reconstruction comparison.} 

\begin{figure*}[t]
    \centering
    \setlength{\tabcolsep}{2.5pt}
    \begin{tabular}{cccc}

        \begin{overpic}[width=0.48\columnwidth]{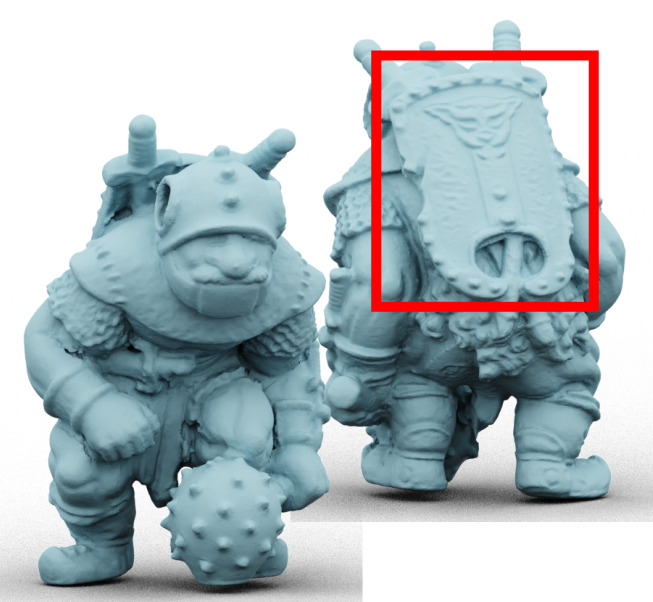} 
            \put ( 0.0, 90.0) {\textbf{(a)}}
        \end{overpic} &
        \includegraphics[width=0.48\columnwidth]{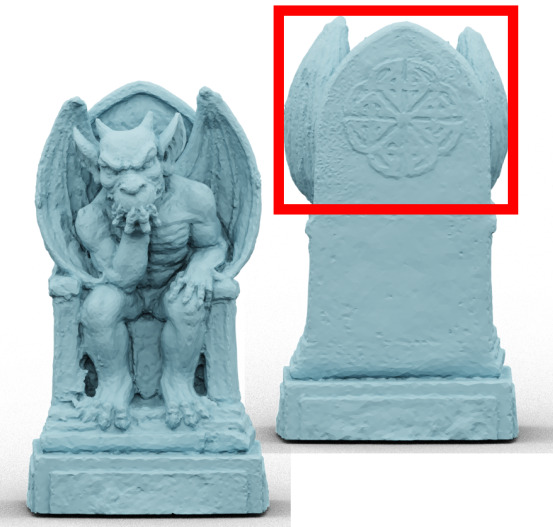} &
        \includegraphics[width=0.48\columnwidth]{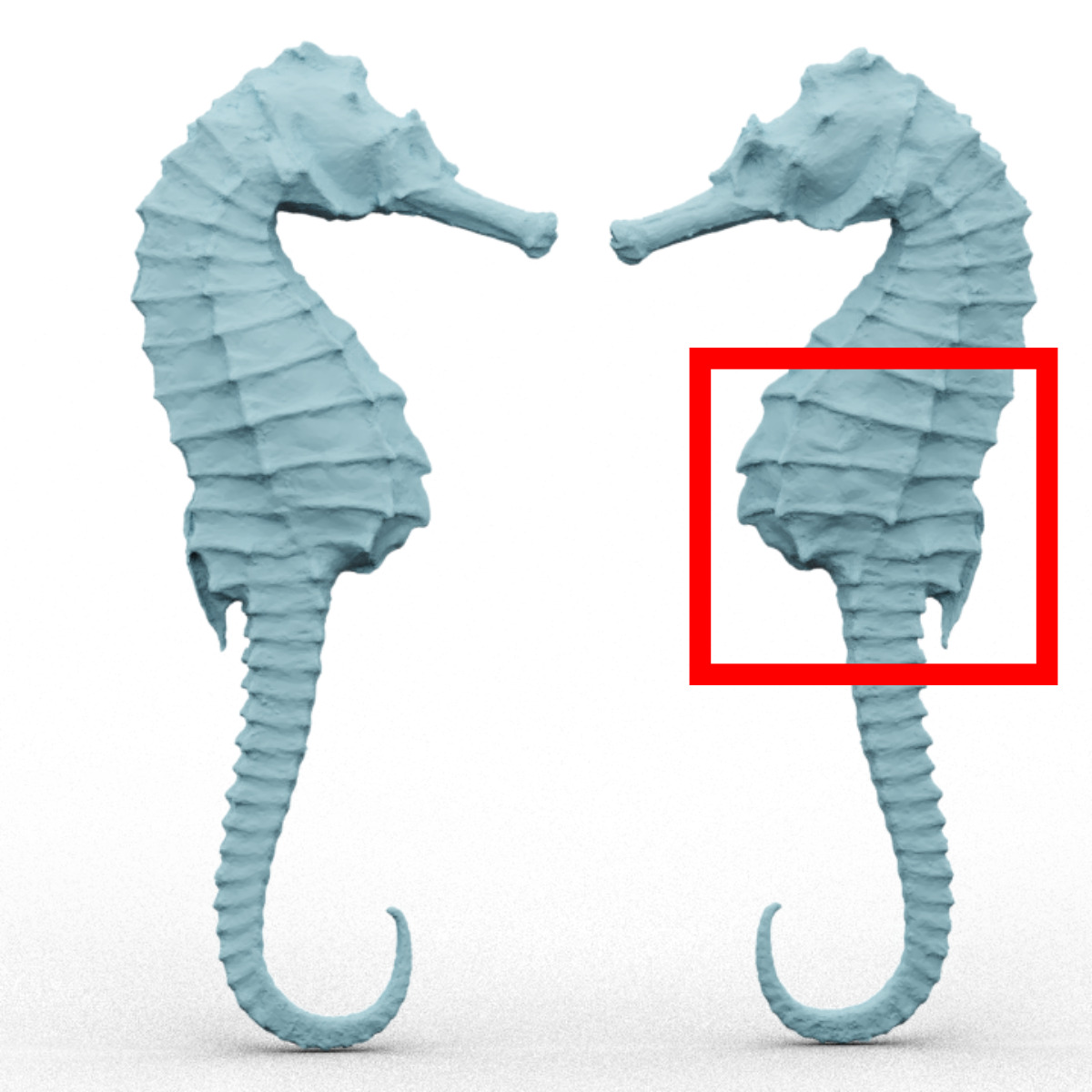} &
        \includegraphics[width=0.48\columnwidth]{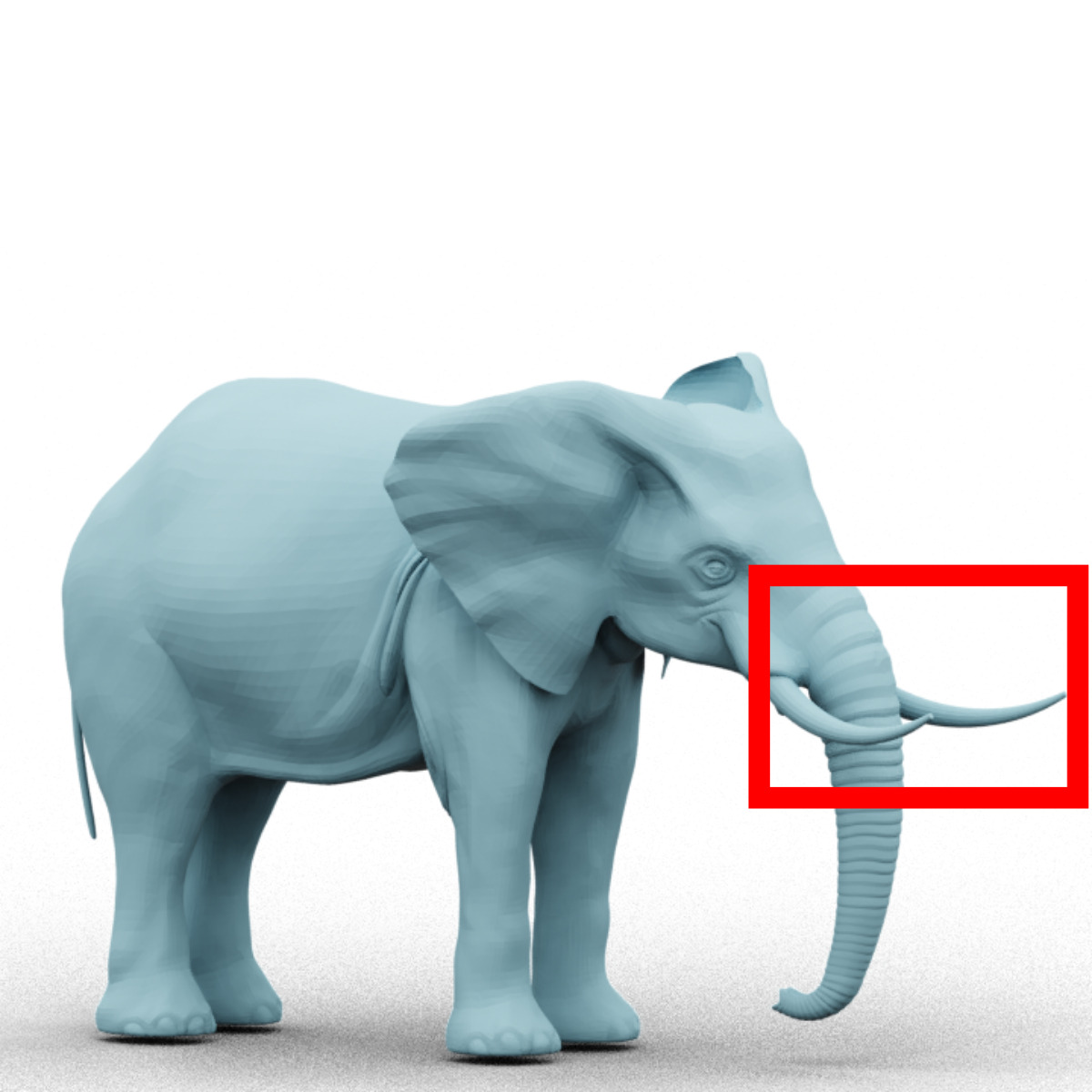} \\
        
        \begin{overpic}[width=0.48\columnwidth]{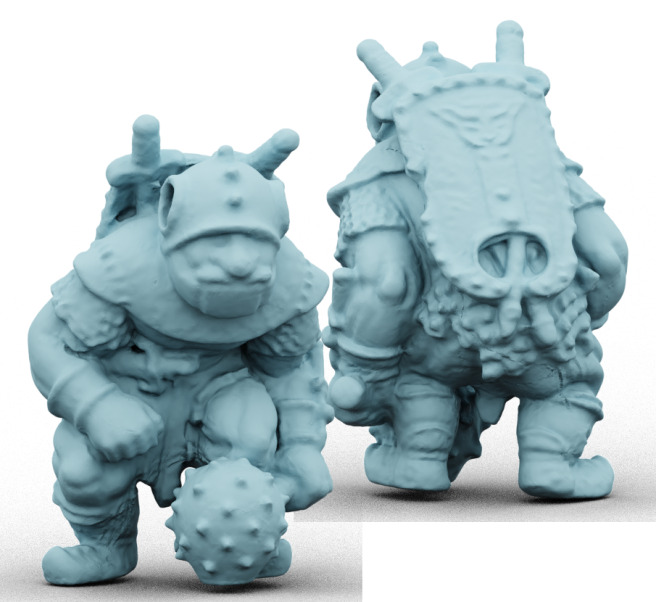} 
            \put ( 0.0, 90.0) {\textbf{(b)}}
        \end{overpic} &
        \includegraphics[width=0.48\columnwidth]{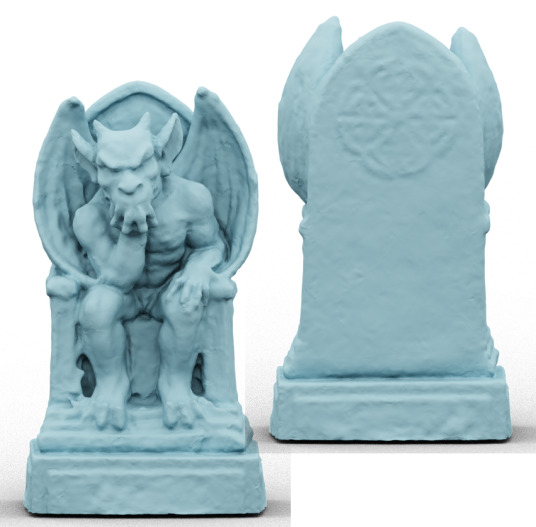} &
        \includegraphics[width=0.48\columnwidth]{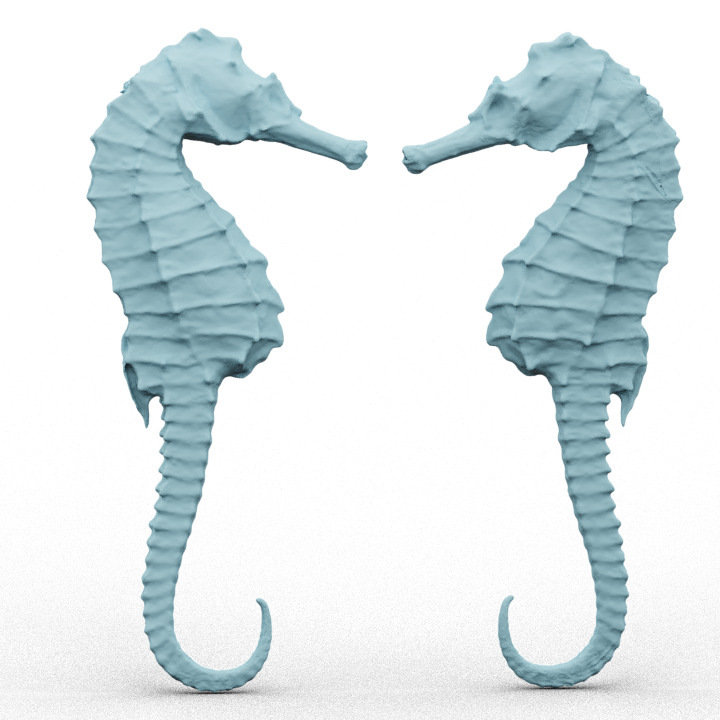} &
        \includegraphics[width=0.48\columnwidth]{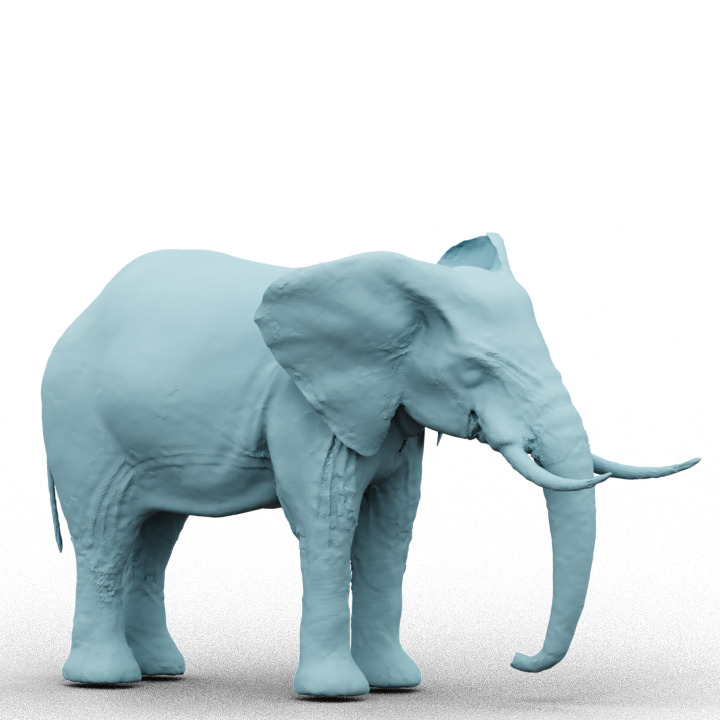} \\
        
        \begin{overpic}[width=0.48\columnwidth]{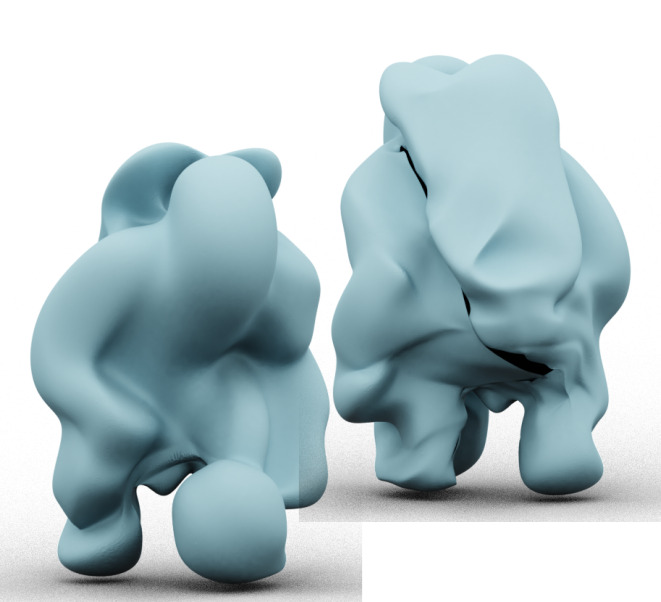} 
            \put ( 0.0, 90.0) {\textbf{(c)}}
        \end{overpic} &
        \includegraphics[width=0.48\columnwidth]{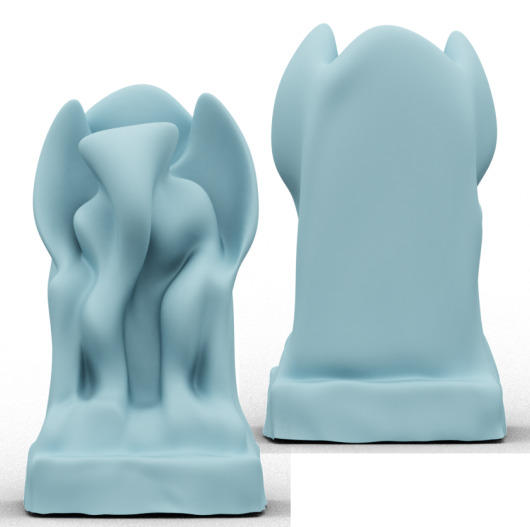} &
        \includegraphics[width=0.48\columnwidth]{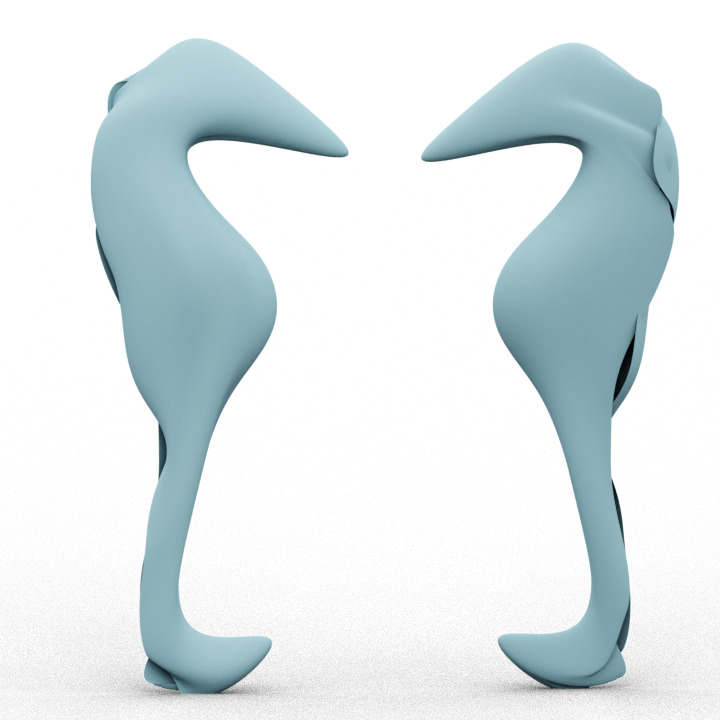} &
        \includegraphics[width=0.48\columnwidth]{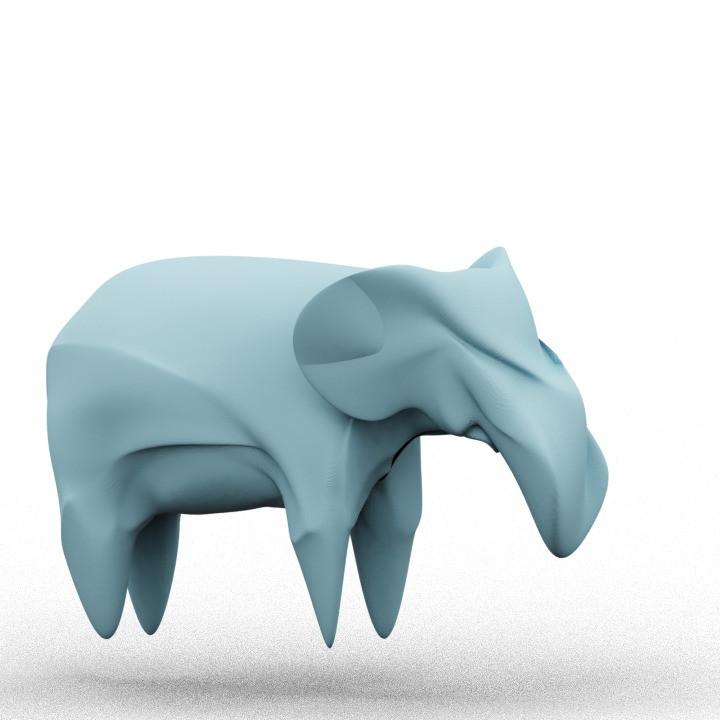} \\

        \begin{overpic}[width=0.48\columnwidth]{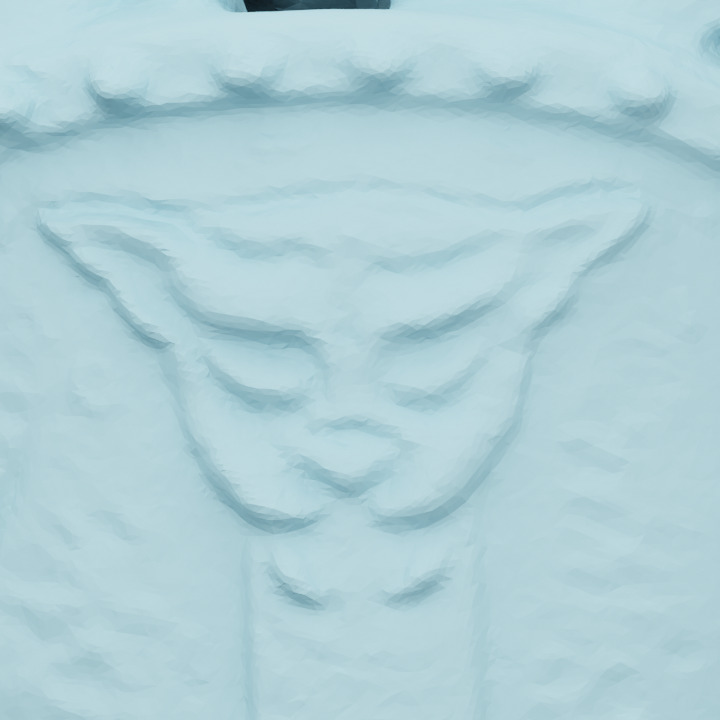} 
            \put ( 0.0, 90.0) {\textbf{(d)}}
        \end{overpic} &
        \includegraphics[width=0.48\columnwidth]{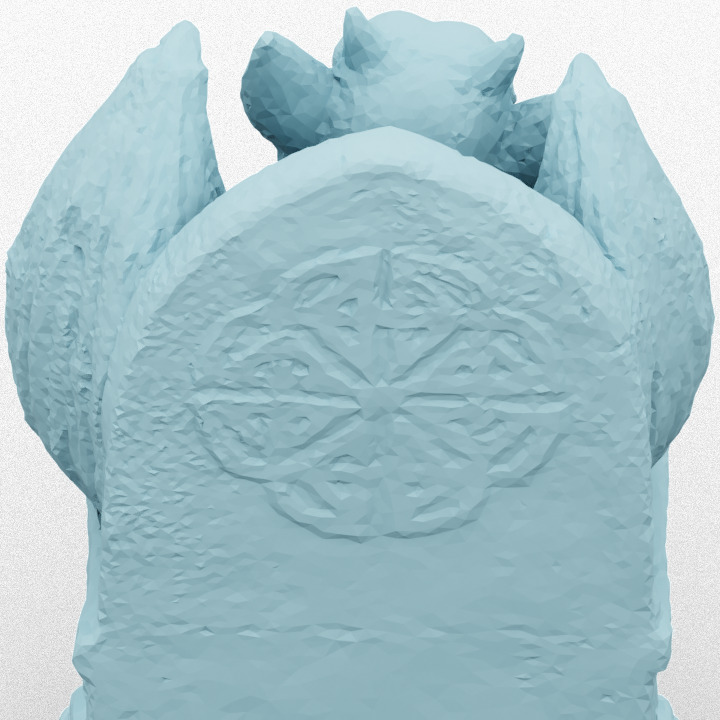} &
        \includegraphics[width=0.48\columnwidth]{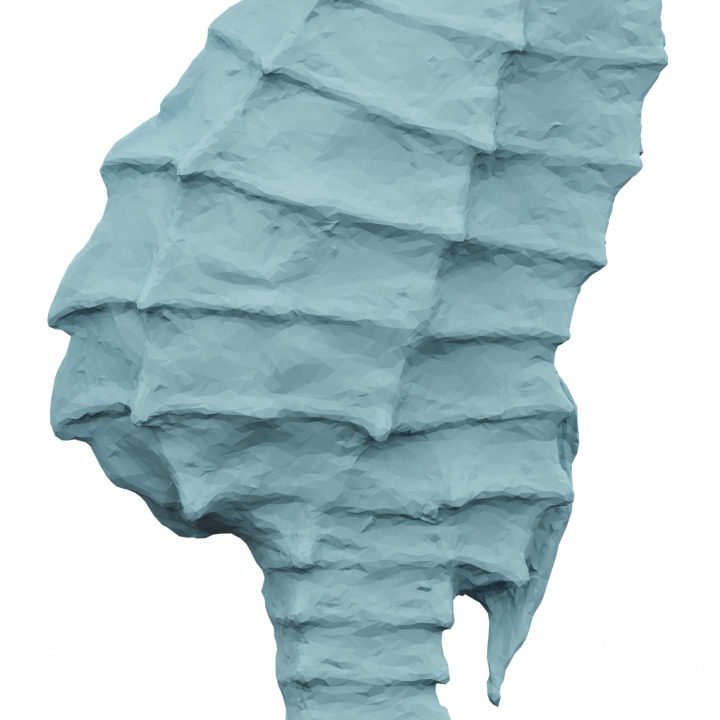} &
        \includegraphics[width=0.48\columnwidth]{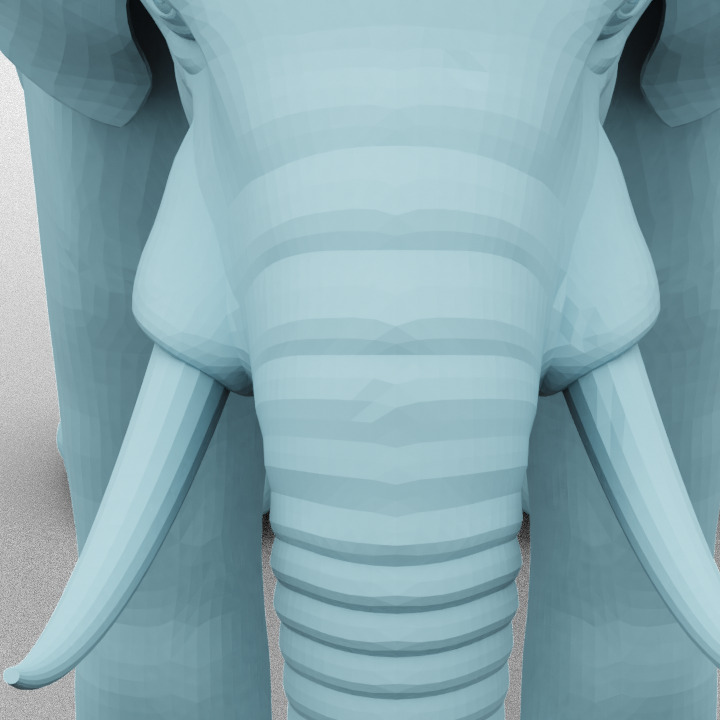} \\

        \begin{overpic}[width=0.48\columnwidth]{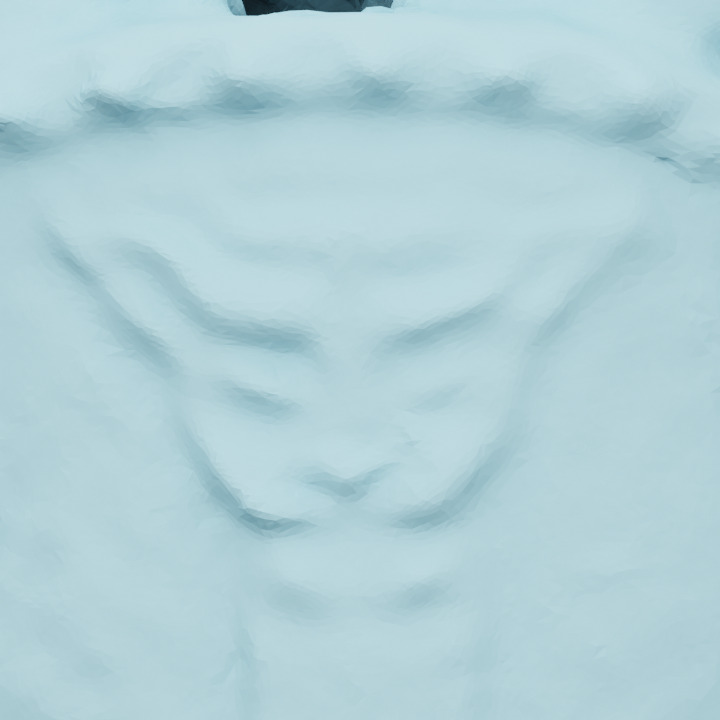} 
            \put ( 0.0, 90.0) {\textbf{(e)}}
        \end{overpic} &
        \includegraphics[width=0.48\columnwidth]{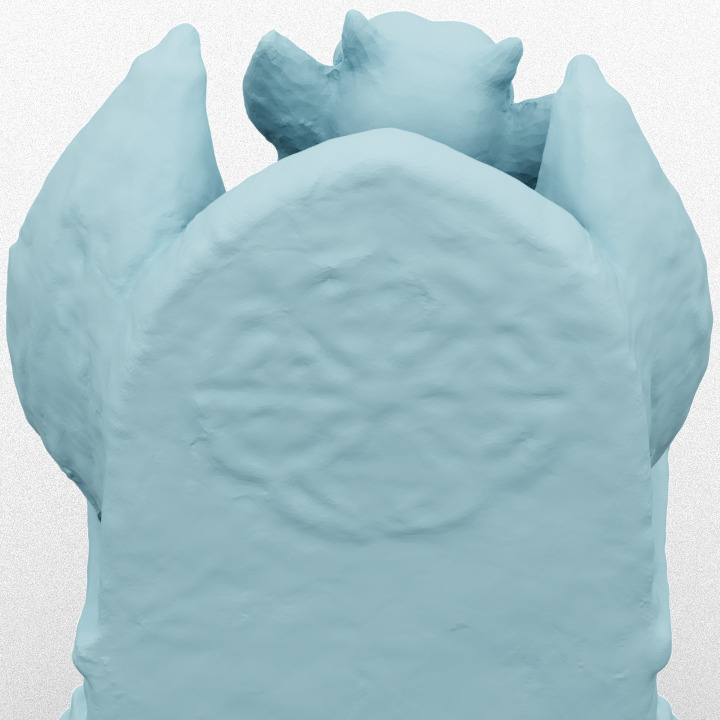} &
        \includegraphics[width=0.48\columnwidth]{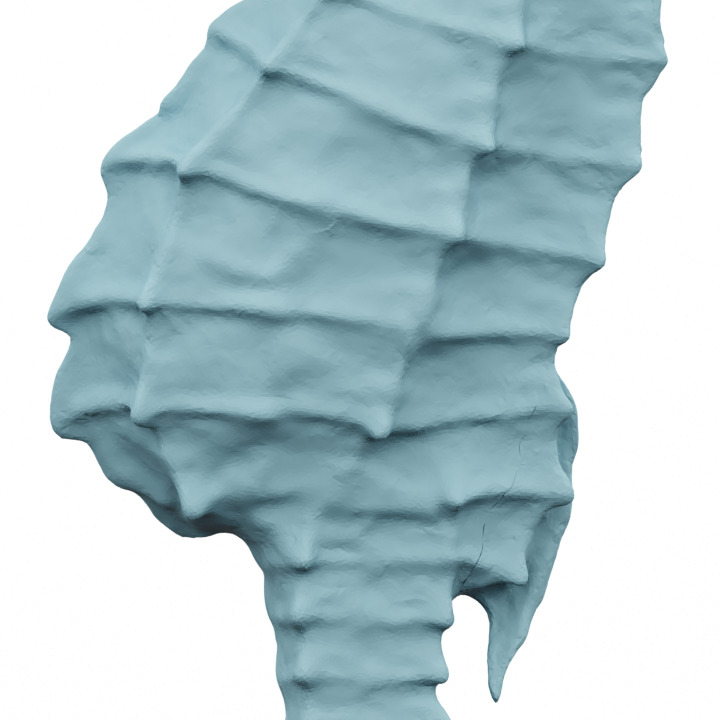} &
        \includegraphics[width=0.48\columnwidth]{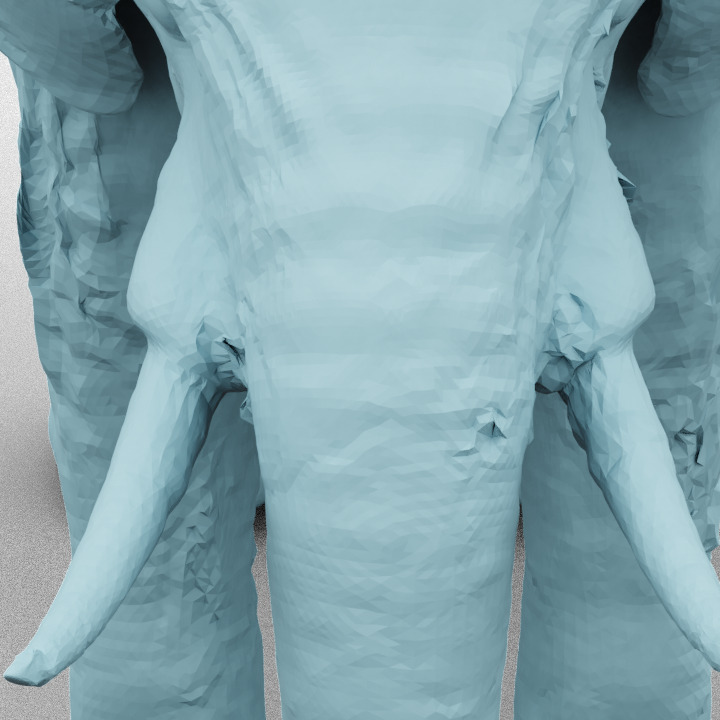} \\
        
        Grog & Gargoyl & Seahorse & Elephant \\
        
    \end{tabular}
    
    \caption{\label{fig:reconstruction_appendix} Representation quality -with~(b) and without~(c) details- of our method, compared with the ground truth model~(a). We limit Neural Convolutional Surfaces to 100K parameters. We show inset zooms~(e)  of our reconstruction for further assessment, with corresponding inset zooms~(d) for ground truth.}
    
\end{figure*}

For all baselines, \cite{martel2021acorn,takikawa2021neural,yifan2021geometryconsistent}, we used authors' implementations. See Figure~\ref{fig:reconstruction_appendix} for qualitative a evaluation of our method, with (b), and without details (c). 

\paragraph{Expressive power.} 
Our construction is readily applicable to any genus, by cutting the mesh to a disk it is possible to reconstruct any surface. See Figure \ref{fig:topology}(a)(b) for reconstruction examples with different genus. 
However, Neural Convolutional Surfaces struggle to represent accurately thin structures. See Figure \ref{fig:topology}(c) for such a thin structure our framework is able to reproduce. 

Finally, the upsampling is fundamental design choice for the CNN. Without upsampling the model is unable to capture details, see Figure \ref{fig:topology}(d).

\section{Architecture Details}
For the model $f_\nu$, we used a 5-layer residual CNN with ReLU non-linearities. The fine MLP $h_\xi$ uses a ReLU non-linearity after each layer except the last, and the coarse MLP $g^c_\phi$ uses Softplus activations. Please refer to Table~\ref{tab:arch_details} for complete architecture details of each model.

\begin{table}[t]
    \centering
    \caption{\label{tab:arch_details} Architecture details used for each shape presented in the paper. }
    
    \setlength{\tabcolsep}{2.5pt}
    \begin{tabular}{r|c|c|c|c}
         & Coarse & Per-patch & CNN $f_\nu$  & Fine \\
         & MLP $g^c_\phi$ & Code $\Omega_i$ & channels & MLP $h_\xi$ \\
         \hline
         \hline
         Armadillo-100K & 128-64-64   & $8 \times 4 \times 4$ & 8 & 16-16 \\
         Bimba-100K     & 128-64      & $8 \times 4 \times 4$ & 8 & 16-16 \\
         Dino-100K      & 64-64       & $8 \times 8 \times 8$ & 8 & 16-16 \\
         Dragon-100K    & 128-64-64   & $6 \times 6 \times 6$ & 8 & 16-16 \\
         Gargoyle-100K  & 64-64-64-64 & $6 \times 4 \times 4$ & 6 & 16-16 \\
         Grog-100K      & 128-64-64   & $8 \times 4 \times 4$ & 8 & 16-16 \\
         Seahorse-100K  & 128-64-64   & $8 \times 8 \times 8$ & 8 & 16-16 \\
         Elephant-100K  & 128-64-64   & $8 \times 6 \times 6$ & 8 & 16-16 \\
         
         \hline
         Armadillo-1M & 128-64-64 & $64 \times 4 \times 4$ & 64 & 16-16 \\
         Bimba-1M     & 128-64    & $64 \times 4 \times 4$ & 64 & 16-16 \\
         Dino-1M      & 64-64     & $64 \times 8 \times 8$ & 64 & 16-16 \\
         Dragon-1M    & 128-64-64 & $66 \times 6 \times 6$ & 64 & 16-16 \\
    \end{tabular}
    
\end{table}

\end{document}